\documentclass[11pt,a4paper]{article}
\usepackage{scrextend}
\usepackage[hyperref]{acl2020}
\usepackage{times}
\usepackage{latexsym}
\usepackage{microtype}
\usepackage{amsfonts, amsmath, amsthm}
\usepackage{breqn}
\usepackage{makecell}
\usepackage{scalerel}

\usepackage{xspace}
\usepackage{url}
\usepackage{relsize}
\usepackage{xfrac}
\usepackage{colortbl,booktabs}
\usepackage{tikz}
\newsavebox{\tempbox}
\usepackage{graphicx}
\usepackage{dblfloatfix}
\definecolor{highlight}{HTML}{90aade}
\newcolumntype{s}{>{\columncolor{highlight}}c}
\usepackage{enumitem}
\usepackage{tipa}
\usepackage{cleveref}
\crefname{section}{\S}{}
\crefname{section}{\S}{\S\S}
\Crefname{section}{\S}{\S\S}    % must define start-of-sentence version explicitly since \S isn't a letter
\crefformat{section}{#2\S#1#3}  % remove space between section symbol and the number, and include \S in the hyperlink
\Crefformat{section}{#2\S#1#3}
\crefname{table}{Table}{Tables}
\crefname{figure}{Figure}{Figures}
\crefname{appendix}{Appendix}{Appendices}
\crefname{enumi}{Caveat}{Caveats}
\crefformat{footnote}{#2\footnotemark[#1]#3}
\usepackage{comment} %to comment sections of a table
\usepackage[hang,flushmargin]{footmisc}
\usepackage{multicol}
\usepackage{multirow}
\usepackage[utf8]{inputenc}
\usepackage{adjustbox}
\usepackage{subcaption}

\newcommand{\corpusname}{VoxClamantis \textsc{\footnotesize{v1.0}}\xspace} 
\newcommand{\numisos}{635\xspace} %num unique iso 639-3 codes
\newcommand{\numrecs}{690\xspace} %num recordings (more than 1 for some iso 639-3)

\newcommand{\ent}{\mathrm{H}}

\usepackage[disable]{todonotes} %toggleme
\makeatletter
\newcommand*\iftodonotes{\if@todonotes@disabled\expandafter\@secondoftwo\else\expandafter\@firstoftwo\fi}  % defines \iftodonotes{<true>}{<false>}, thanks to https://tex.stackexchange.com/questions/126559/conditional-based-on-packageoption
\makeatother

\newcommand{\note}[4][]{\todo[author=#2,color=#3,size=\scriptsize,fancyline,caption={},#1]{#4}} % default note settings, used by macros below.
\newcommand{\response}[1]{\vspace{3pt}\hrule\vspace{3pt}\textbf{#1:} } % insert \response{myname} within someone else's todonotetphone
\newlength{\extramargin}
\usepackage{calc}
\iftodonotes{\usepackage[paperwidth=\paperwidth+\extramargin*2,marginparwidth=\marginparwidth+\extramargin,width=\textwidth,height=\textheight]{geometry}}{}

\newcommand{\jason}[2][]{\note[#1]{jason}{green!40}{#2}}

\definecolor{gr}{RGB}{169,169,169}

\newcommand{\phon}[1]{/\textipa{#1}/}
\newcommand{\graph}[1]{$<$\texttt{#1}$>$}

\everypar{\looseness=-1} %this command does not work and I do not know how to make it work. so I am adding this to every paragraph the manual way -liz

\newcommand{\link}{\normalsize{\url{https://voxclamantisproject.github.io}}}

\newcommand{\jhu}{\textrm{\normalfont \textipa{C}}}
\newcommand{\cmu}{\textrm{\normalfont \textbeltl}}
\newcommand{\york}{\textrm{\normalfont \textipa{7}}}
\newcommand{\ucam}{\textrm{\normalfont \textipa{Z}}}
\newcommand{\ethz}{\textrm{\normalfont \textipa{D}}}

\aclfinalcopy % Uncomment this line for the final submission
\def\aclpaperid{1176} %  Enter the acl Paper ID here
\title{A Corpus for Large-Scale Phonetic Typology}

\author{Elizabeth Salesky$^\jhu$~\;~Eleanor Chodroff$^\york$~\;~Tiago Pimentel$^\ucam$~\;~ Matthew Wiesner$^\jhu$ \\
\textbf{Ryan Cotterell}$^{\ucam,\ethz}$~\;~\textbf{Alan W Black}$^\cmu$~\;~\textbf{Jason Eisner}$^\jhu$ \\
  $^\jhu$Johns Hopkins University~\;~ $^\york$University of York~\;~ \\
  $^\ucam$University of Cambridge~\;~$^\ethz$ETH Z{\"u}rich~\;~ $^\cmu$Carnegie Mellon University \\
  \texttt{esalesky@jhu.edu}~\;~\texttt{eleanor.chodroff@york.ac.uk}
}

\begin{document}
\maketitle

%----------------
\begin{abstract}
A major hurdle in data-driven research on typology is having sufficient data in many languages to draw meaningful conclusions. 
We present \corpusname, the first large-scale corpus for phonetic typology, with aligned segments and estimated phoneme-level labels in \numrecs readings spanning \numisos languages, along with acoustic-phonetic measures of vowels and sibilants.
Access to such data can greatly facilitate investigation of phonetic typology at a large scale and across many languages.
However, it is non-trivial and computationally intensive to obtain such alignments for hundreds of languages, many of which have few to no resources presently available. 
We describe the methodology to create our corpus, discuss caveats with current methods and their impact on the utility of this data, and illustrate possible research directions through a series of case studies on the 48 highest-quality readings. Our corpus and scripts are publicly available for non-commercial use at \link.
\end{abstract}

\section{Introduction}

Understanding the range and limits of cross-linguistic variation is fundamental to the scientific study of language. 
In speech and particularly phonetic typology, this involves exploring potentially universal tendencies that shape sound systems and govern phonetic structure. 
Such investigation requires access to large amounts of cross-linguistic data.
Previous cross-linguistic phonetic studies have been limited to a small number of languages with available data \citep{Disner1983, ChoLadefoged1999}, or have relied on previously reported measures from many studies \citep{WhalenLevitt1995,BeckerKristal2010,GordonRoettger2017,ChodroffEtAl2019}. 
Existing multilingual speech corpora have similar restrictions, with data too limited for many tasks \cite{Engstrand1988, UCLAPhoneticsLab} or approximately 20 to 30 recorded languages \citep{ardila2019common, babel2011iarpa, schultz2002globalphone}.\looseness=-1

%-- liz: putting here to force latex placement in top right -- https://twitter.com/aclanthology/status/1179582815840505856
\begin{figure}[t]
 \centering
 \includegraphics[width=\columnwidth]{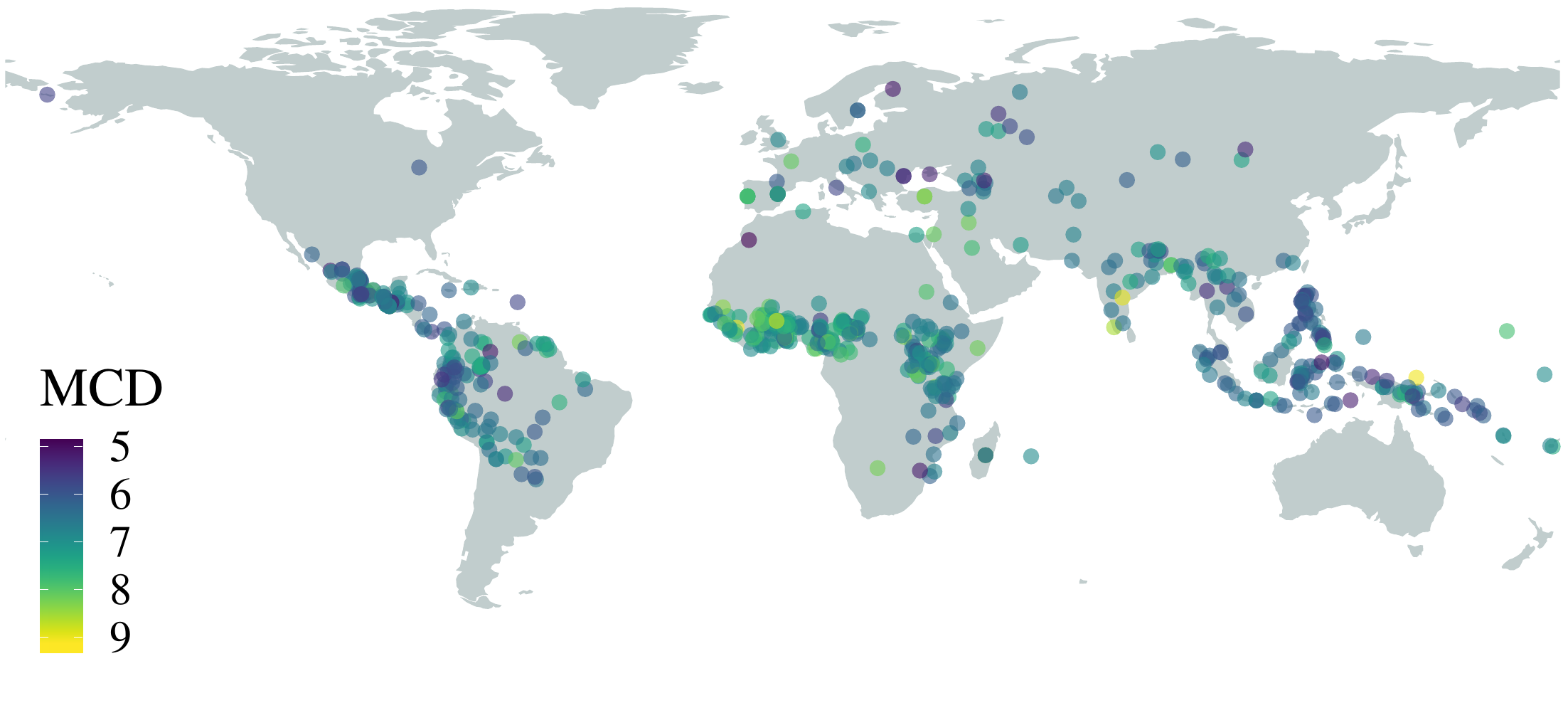}
 \vspace{-2em}
 \caption{The \numisos languages of our corpus geo-located with mean Mel Cepstral Distortion (MCD) scores.}
 \label{fig:mcd-map}
\end{figure}

The recently developed CMU Wilderness corpus \cite{black2019cmu} constitutes an exception to this rule with over 600 languages. 
This makes it the largest and most typologically diverse speech corpus to date. 
In addition to its coverage, the CMU Wilderness corpus is unique in two additional aspects: cleanly recorded, read speech exists for all languages in the corpus, and the same content (modulo translation) exists across all languages.\looseness=-1 

However, this massively multilingual speech corpus is challenging to work with directly.
Copyright, computational restrictions, and sheer size limit its accessibility. 
Due to copyright restrictions, the audio cannot be directly downloaded with the sentence and phoneme alignments. 
A researcher would need to download original audio MP3 and text through links to \url{bible.is}, then segment these with speech-to-text sentence alignments distributed in \citet{black2019cmu}.\footnote{The stability of the links and recording IDs is also questionable. Since the release of \citet{black2019cmu}, many of the links have already changed, along with a few of the IDs. We have begun identifying these discrepancies, and plan to flag these in a future release.}
For phonetic research, subsequently identifying examples of specific phonetic segments in the audio is also a near-essential step for extracting relevant acoustic-phonetic measurements. 
Carrying out this derivative step has allowed us to release a stable-access collection of token-level acoustic-phonetic measures to enable further research.\looseness=-1

Obtaining such measurements requires several processing steps: estimating pronunciations,
aligning them to the text,
evaluating alignment quality, and finally, extracting phonetic measures.  
This work is further complicated by the fact that, for a sizable number of these languages, no linguistic resources currently exist (e.g., language-specific pronunciation lexicons).
We adapt speech processing methods based on \citet{black2019cmu} to accomplish these tasks, though not without noise: in \cref{sec:caveats}, we identify three significant caveats when attempting to use our extended corpus for large-scale phonetic studies.\looseness=-1 

We release a comprehensive set of standoff markup of over 400 million labeled segments of continuous speech.\footnote{For some languages, we provide multiple versions of the markup based on different methods of predicting the pronunciation and generating time alignments (\cref{sec:phone_alignments}).}
For each segment, we provide an estimated phoneme-level label from the X-SAMPA alphabet, the preceding and following labels, and the start position and duration in the audio.
Vowels are supplemented with formant measurements, and sibilants with standard measures of spectral shape.
\looseness=-1

We present a series of targeted case studies illustrating the utility of our corpus for large-scale phonetic typology. 
These studies are motivated by potentially universal principles posited to govern phonetic variation: \textbf{phonetic dispersion} and \textbf{phonetic uniformity}.
Our studies both replicate known results in the phonetics literature and also present novel findings.
Importantly, these studies investigate current methodology as well as questions of interest to phonetic typology at a large scale.\looseness=-1 
\begin{figure}[h]
 \centering
 \includegraphics[width=\columnwidth]{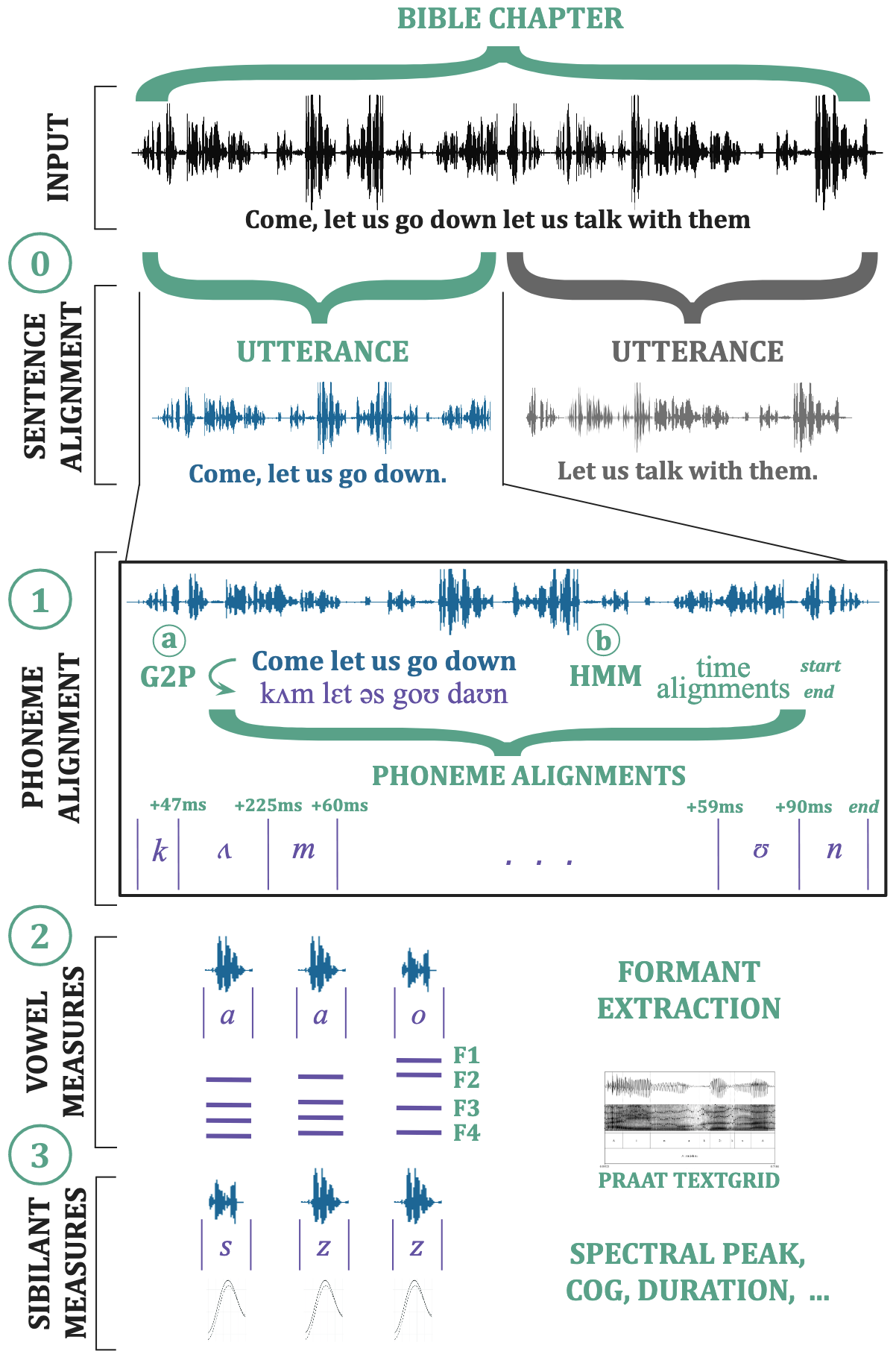}
 \caption{The extraction process for the measurements released in \corpusname.}
 \label{fig:extraction-process}
\end{figure}

%-----------------------------------------
\section{Original Speech}
\label{sec:cmu}

The CMU Wilderness corpus \citep{black2019cmu} consists of recorded readings of the New Testament of the Bible in many languages and dialects. 
Following the New Testament structure, these data are broken into 27 books, each with a variable number of chapters between 1 and 25.
Bible chapters contain standardized verses (approximately sentence-level segments); 
however, the speech is originally split only by chapter. 
Each chapter has an average of 13 minutes of speech for a total of $\approx$20 hours of speech and text per language. 
These recordings are clean, read speech with a sampling rate of 16 kHz. 
In most languages, they are non-dramatic readings with a single speaker; in some, they are dramatic multi-speaker readings with additive music.\footnote{Information about the recordings available can be found at \url{https://www.faithcomesbyhearing.com/mission/recordings}}
The release from \citet{black2019cmu} includes several resources for processing the corpus: scripts to download the original source data from \url{bible.is}, `lexicons' created using grapheme-to-phoneme (G2P) conversion, and scripts to apply their generated sentence alignments, which facilitates downstream language processing tasks, including phoneme alignment.\looseness=-1

%-----------------------------------------------
\section{The \corpusname Corpus}

Our \corpusname corpus is derived from \numrecs audio readings of the New Testament of the Bible\footnote{Nine of the readings from \citet{black2019cmu} could not be aligned.} in \numisos languages.\footnote{We specify number of distinct languages by the number of distinct ISO 639-3 codes, which may not distinguish dialects.\looseness=-1 } 
We mark estimated speech segments labeled with phonemic labels, and phonetic measures for the tokens that are vowels or sibilants. 
The extraction process is diagrammed in \cref{fig:extraction-process}. 
In the sections below, we detail our procedures for extracting labeled audio segments and their phonetic measures, in both high- and low-resource languages. 
We then outline important caveats to keep in mind when using this corpus.

%------------------------------------
\subsection{Extracting Phoneme Alignments}
\label{sec:phone_alignments}

We use a multi-pronged forced alignment strategy to balance broad language coverage (\cref{sec:unitran}) with utilization of existing high-quality resources (\cref{sec:hrl}).  We assess the quality of our approaches in \cref{sec:qualitymeasures}. 
We release the stand-off markup for our final alignments as both text files and Praat TextGrids \citep{praat}.\footnote{Corresponding audio will need to be downloaded from source and split by utterance using scripts from \citet{black2019cmu}.}\looseness=-1 

Using scripts and estimated boundaries from \citet{black2019cmu}, we first download and convert the audio MP3s to waveforms, and cut the audio and text into `sentences' (hereafter called `utterances' as they are not necessarily sentences). 
This step creates shorter-length speech samples to facilitate forced alignment; utterance boundaries do not change through our processing. 

To extract labeled segments, we first require pronunciations for each utterance. 
A pronunciation is predicted from the text alone using some grapheme-to-phoneme (G2P) method.  Each word's predicted pronunciation is a sequence of categorical labels, which are `phoneme-level' in the sense that they are usually intended to distinguish the words of the language. We then align this predicted sequence of `phonemes' to the corresponding audio. 

%------------------------------------
\subsubsection{All Languages} 
\label{sec:unitran}

Most of our languages have neither existing pronunciation lexicons nor G2P resources. 
To provide coverage for all languages, we generate pronunciations using the simple `universal' G2P system Unitran (\citealp{qian2010python}, as extended by \citealp{black2019cmu}), which deterministically expands each grapheme to a fixed sequence of phones in the Extended Speech Assessment Methods Phonetic Alphabet (X-SAMPA) \citep{wells1995computer}. 
This naive process is error-prone for languages with opaque orthographies, as we show in \cref{sec:qualitymeasures} below and discuss further in \cref{sec:caveats} (\cref{caveat:phonelabels}).
Even so, it provides a starting point for 
exploring low-resource languages: after some manual inspection, a linguist may be able to correct the labels in a given language by a combination of manual and automatic methods.\looseness=-1

For each reading, to align the pronunciation strings to the audio, we fit a generative acoustic model designed for this purpose: 
specifically, eHMM \citep{prahallad2006sub} as implemented in Festvox \citep{anumanchipalli2011festvox} to run full Baum--Welch from a flat start for 15 to 30 iterations until the mean mel cepstral distortion score (see \cref{sec:qualitymeasures}) converges.  
Baum-Welch does not change the predicted phoneme labels, but obtains a language-specific, reading-specific, contextual (triphone) acoustic model for each phoneme type in the language. 
We then use Viterbi alignment to identify an audio segment for each phoneme token.\looseness=-1

%--------------------------------------
\subsubsection{High-Resource Languages}
\label{sec:hrl}

A subset of the languages in our corpus are supported by existing pronunciation resources.  
Two such resources are Epitran \citep{epitran2018mortensen}, a G2P tool based on language-specific rules, available in both IPA and X-SAMPA, and WikiPron \citep{wikipron}, a collection of crowd-sourced pronunciations scraped from Wiktionary. These are mapped from IPA to X-SAMPA for label consistency across our corpus. 
Epitran covers 29 of our languages (39 readings), while WikiPron's `phonemic' annotations\footnote{WikiPron annotations are available at both the phonemic and phonetic level, with a greater number of phonemic annotations, which we use here.} provide partial coverage of 13 additional languages (18 readings).
We use Epitran for languages with regular orthographies where it provides high-quality support, and WikiPron for other languages covered by WikiPron annotations.
While Unitran and Epitran provide a single pronunciation for a word from the orthography, WikiPron may include multiple pronunciations. 
In such cases, Viterbi alignment (see below) chooses the pronunciation of each token that best fits the audio.\looseness=-1  

%-----------------------------------
\begin{table*}[th]
\footnotesize
\begin{adjustbox}{width=\linewidth}
\begin{tabular}{lrrrrrrrrrrrrrrr}
\toprule
\textbf{ISO 639-3} & \textbf{tpi} & \textbf{ron} & \textbf{azj} & \textbf{msa} & \textbf{ceb} & \textbf{tur} & \textbf{tgl} & \textbf{spa} & \textbf{ilo} & \textbf{rus} & \textbf{hau} & \textbf{ind} & \textbf{tgk} & \textbf{jav} & \textbf{kaz} \\
\cmidrule(lr){2-2} \cmidrule(lr){3-3} \cmidrule(lr){4-4} \cmidrule(lr){5-5} \cmidrule(lr){6-6} \cmidrule(lr){7-7} \cmidrule(lr){8-8} \cmidrule(lr){9-9} \cmidrule(lr){10-10} \cmidrule(lr){11-11} \cmidrule(lr){12-12} \cmidrule(lr){13-13} \cmidrule(lr){14-14} \cmidrule(lr){15-15} \cmidrule(lr){16-16}
\# Types              & 1398         & 9746         & 18490        & 7612         & 8531         & 21545        & 9124         & 11779        & 15063        & 16523        & 4938         & 5814         & 12502        & 10690        & 20502        \\
Unitran PER        & 18.4         & 21.3         & 26.9         & 30.1         & 30.1         & 31.2         & 34.4         & 34.4         & 35.0         & 37.4         & 37.6         & 38.8         & 39.8         & 49.9         & 46.8         \\
\cmidrule(lr){2-2} \cmidrule(lr){3-3} \cmidrule(lr){4-4} \cmidrule(lr){5-5} \cmidrule(lr){6-6} \cmidrule(lr){7-7} \cmidrule(lr){8-8} \cmidrule(lr){9-9} \cmidrule(lr){10-10} \cmidrule(lr){11-11} \cmidrule(lr){12-12} \cmidrule(lr){13-13} \cmidrule(lr){14-14} \cmidrule(lr){15-15} \cmidrule(lr){16-16}
\# Tokens    & 291k & 169k & 125k & 157k & 190k & 125k & 185k & 168k & 169k & 130k & 201k & 170k & 159k & 177k & 142k \\
Weighted PER & 20.1 & 21.3 & 26.1 & 31.1 & 35.9 & 28.5 & 40.1 & 32.6 & 32.7 & 36.8 & 36.7 & 40.5 & 38.8 & 54.1 & 47.7 \\
\midrule
\textbf{ISO 639-3} & \textbf{swe} & \textbf{kmr} & \textbf{som} & \textbf{tir} & \textbf{pol} & \textbf{hae} & \textbf{vie} & \textbf{tha} & \textbf{lao} & \textbf{ben} & \textbf{tel} & \textbf{hin} & \textbf{mar} & \textbf{tam} & \textbf{}    \\
\cmidrule(lr){2-2} \cmidrule(lr){3-3} \cmidrule(lr){4-4} \cmidrule(lr){5-5} \cmidrule(lr){6-6} \cmidrule(lr){7-7} \cmidrule(lr){8-8} \cmidrule(lr){9-9} \cmidrule(lr){10-10} \cmidrule(lr){11-11} \cmidrule(lr){12-12} \cmidrule(lr){13-13} \cmidrule(lr){14-14} \cmidrule(lr){15-15} 
\# Types              & 8610         & 8127         & 14375        & 22188        & 18681        & 15935        & 2757         & 23338        & 31334        & 8075         & 23477        & 7722         & 17839        & 31642        &              \\
Unitran PER        & 46.9         & 54.3         & 54.6         & 57.8         & 67.1         & 67.3         & 73.8         & 80.3         & 89.1         & 90.0         & 90.3         & 95.7         & 97.8         & 100.5        &             \\
\cmidrule(lr){2-2} \cmidrule(lr){3-3} \cmidrule(lr){4-4} \cmidrule(lr){5-5} \cmidrule(lr){6-6} \cmidrule(lr){7-7} \cmidrule(lr){8-8} \cmidrule(lr){9-9} \cmidrule(lr){10-10} \cmidrule(lr){11-11} \cmidrule(lr){12-12} \cmidrule(lr){13-13} \cmidrule(lr){14-14} \cmidrule(lr){15-15} 
\# Tokens    & 165k & 176k & 156k & 121k & 141k & 164k & 211k &  26k &  36k & 173k & 124k & 191k & 159k & 139k \\
Weighted PER & 49.5 & 53.9 & 56.0 & 57.4 & 66.8 & 64.8 & 80.6 & 80.4 & 89.4 & 86.2 & 88.3 & 91.3 & 97.8 & 102.1 \\
\bottomrule
\end{tabular}
\end{adjustbox}
\caption{Phoneme Error Rate (PER) for Unitran treating Epitran as ground-truth. `Types' and `Tokens' numbers reflect the number of unique word types and word tokens in each reading. We report PER calculated using word types for calibration with other work, as well as frequency-weighted PER reflecting occurrences in our corpus.}
\label{tab:PER}
\end{table*}
%-----------------------------------

For most languages covered by WikiPron, most of our corpus words are out-of-vocabulary, as they do not yet have user-submitted pronunciations on Wiktionary. 
We train G2P models on WikiPron annotations to provide pronunciations for these words. \jason{would be a neat trick to provide \emph{multiple} plausible pronunciations and choose among them as mentioned above; that is basically the key to unsupervised learning\response{matthew} I considered this actually and this is one way that people bootstrap lexicon learning in ASR from a seed lexicon. To evaluate whether we actually chose better pronunciations though, we would need to evaluate some downstream task (like ASR), since we have no ground truth. This is actually easy to do, but there wasn't enough time here for it. \response{Jason} Cool. Do you have a ref?  might want to mention in future work \cite{lu2013acoustic, zhang2017acoustic}}
Specifically, we use the WFST-based tool Phonetisaurus \citep{novak2016phonetisaurus}. 
Model hyperparameters are tuned on 3 WikiPron languages from SIGMORPHON 2020 \citep{gorman2020sigmorphon} (see \cref{app:wikipron_acc} for details). 
In general, for languages that are not easily supported by Epitran-style G2P \emph{rules}, training a G2P \emph{model} on sufficiently many high-quality annotations may be more accurate.\looseness=-1

We align the speech with the high-quality labels using a multilingual ASR model \citep[see][]{wiesner2019asru}.
The model is trained in Kaldi \citep{povey2011kaldi} on 300 hours of data from the IARPA BABEL corpora (21 languages), a subset of Wall Street Journal (English), the Hub4 Spanish Broadcast news (Spanish), and a subset of the Voxforge corpus (Russian and French). 
These languages use a shared X-SAMPA phoneme label set which has high coverage of the labels of our corpus.\looseness=-1

Our use of a pretrained multilingual model here contrasts with \cref{sec:unitran}, where we had to train reading-specific acoustic models to deal with the fact that the same Unitran phoneme label may refer to quite different phonemes in different languages (see \cref{sec:caveats}). 
We did not fine-tune our multilingual model to each language, as the cross-lingual ASR performance in previous work \citep{wiesner2019asru} suggests that this model is sufficient for producing phoneme-level alignments.\looseness=-1
\jason{I'd prefer to restore the comment about consistent placement of phoneme boundaries.  I don't think it invalidates Eleanor's experiments as she fears; rather, it improves them by comparing apples with apples. I'm having trouble seeing what could go wrong.  (But if something could go wrong, we should admit it rather than deleting it :-P) \response{liz} we actually deleted it not because of eleanor's concern but matthew's suggestion that we shouldn't claim more than we know -- we know from ASR models that these segments transferred very well without fine-tuning, but without directly comparing he didn't want to claim anything about e.g. boundary comparability etc
\response{eleanor} didn't mean to imply we should delete it to hide something; it just seemed speculative and at that point, I wasn't reading it as something positive for my analyses, per se, just as something that could prompt someone to go: but hey, you built (will build) the correlations in (which yes, if we did, we should be honest, but I was mainly getting speculation from the writing \response{jason} Fair enough that it's speculative; but it is a possible effect of using a multilingual model (and a possible reason to prefer one) and so seemed worth alerting the reader to}

%-----------------------------------------
\subsubsection{Quality Measures}
\label{sec:qualitymeasures}
%---------

Automatically generated phoneme-level labels and alignments inherently have some amount of noise, and this is particularly true for low-resource languages. 
The noise level is difficult to assess without gold-labeled corpora for either modeling or assessment. 
However, for the high-resource languages, we can evaluate Unitran against Epitran and WikiPron, pretending that the latter are ground truth.  For example, \cref{tab:PER} shows Unitran's phoneme error rates relative to Epitran.  \Cref{app:ali_hists} gives several more detailed analyses with examples of individual phonemes.\looseness=-1

Unitran pronunciations may have acceptable phoneme error rates for languages with transparent orthographies and one-to-one grapheme-to-phoneme mappings.  Alas, without these conditions they prove to be highly inaccurate.

%--PER grain of salt--
That said, evaluating Unitran labels against Epitran or WikiPron may be unfair to Unitran, 
since some discrepancies are arguably not errors but mere differences in annotation granularity. 
For example, the `phonemic' annotations in WikiPron are sometimes surprisingly fine-grained: WikiPron frequently uses /\textsubbridge{\textipa{t}}/ in Cebuano where Unitran only uses \phon{t}, though these refer to the same phoneme. These tokens are scored as incorrect. 
Moreover, there can be simple systematic errors: Unitran always maps grapheme \graph{a} to label \phon{A}, but in Tagalog, all such tokens should be \phon{a}.  
Such errors can often be fixed by remapping the Unitran labels, which in these cases would reduce PER from 30.1 to 6.8 (Cebuano) and from 34.4 to 7.8 (Tagalog). 
Such rules are not always this straightforward and should be created on a language-specific basis; we encourage rules created for languages outside of current Epitran support to be contributed back to the Epitran project.\looseness=-1 

For those languages where we train a G2P system on WikiPron, we compute the PER of the G2P system on held-out WikiPron entries treated as ground truth.  The results  (\cref{app:wikipron_acc}) range from excellent to mediocre.

We care less about the pronunciations themselves than about the segments that we extract by aligning these pronunciations to the audio.  
For high-resource languages, we can again compare the segments extracted by Unitran to the higher-quality ones extracted with better pronunciations.  For each Unitran token, we evaluate its label and temporal boundaries against the high-quality token that is closest in the audio, as measured by the temporal distance between their midpoints (\cref{app:ali_hists}).\looseness=-1

Finally, the segmentation of speech and text into corresponding utterances is not perfect. We use the utterance alignments generated by \citet{black2019cmu}, in which the text and audio versions of a putative utterance may have only partial overlap.
Indeed, \citet{black2019cmu} sometimes failed to align the Unitran pronunciation to the audio at all, and discarded these utterances.  For each remaining utterance, he assessed the match quality using Mel Cepstral Distortion (MCD)---which is commonly used to evaluate synthesized spoken utterances \citep{kominek2008synthesizer}---between the original audio and a resynthesized version of the audio based on the aligned pronunciation.  Each segment's audio was resynthesized given the segment's phoneme label and the preceding and following phonemes, in a way that preserves its duration, using CLUSTERGEN \cite{black2006clustergen} with the same reading-specific eHMM model that we used for alignment. 
We distribute \citeauthor{black2019cmu}'s per-utterance MCD scores with our corpus, and show the average score for each language in \cref{app:lang_list}.  In some readings, the MCD scores are consistently poor.\looseness=-1

%-----------------------------------------
\subsection{Phonetic measures}\label{sec:phonetic-measures}

Using the phoneme-level alignments described in \cref{sec:phone_alignments}, we automatically extract several standard acoustic-phonetic measures of vowels and sibilant fricatives that correlate with aspects of their articulation and abstract representation.\looseness=-1

%------------------------------------
\subsubsection{Vowel measures}
\label{sec:vowels}

\begin{figure}[t]
 \centering
 \includegraphics[width=0.75\columnwidth]{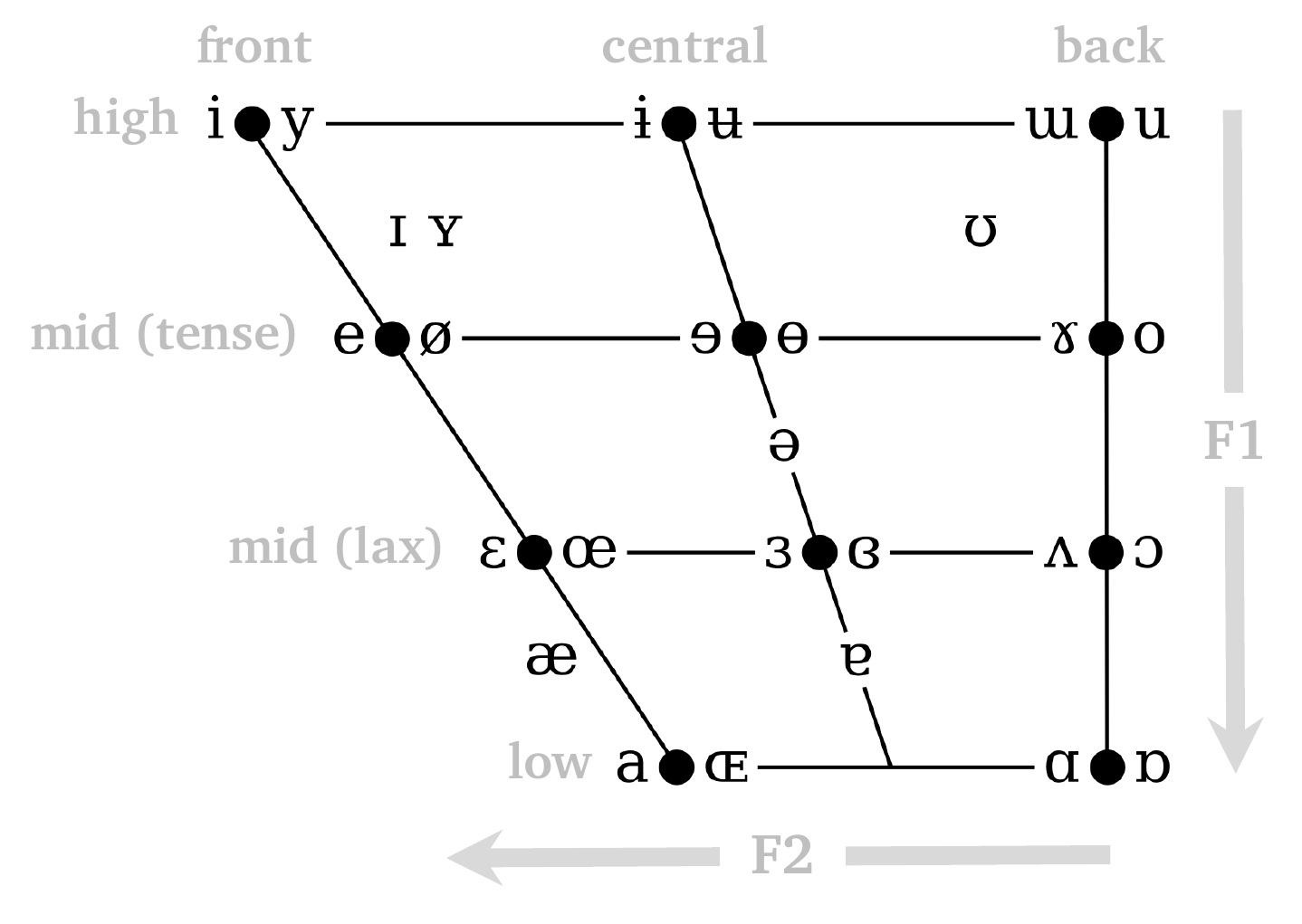}
 \caption{Vowel Chart}
\label{fig:vowel_chart}
\end{figure}

Standard phonetic measurements of vowels include the formant frequencies and duration information. 
Formants are concentrations of acoustic energy at frequencies reflecting resonance points in the vocal tract during vowel production \cite{LadefogedJohnson2014}.
The lowest two formants, F1 and F2, are considered diagnostic of vowel category identity and approximate tongue body height (F1) and backness (F2) during vowel production (\cref{fig:vowel_chart}). 
F3 correlates with finer-grained aspects of vowel production such as rhoticity (\phon{r}-coloring), lip rounding, and nasality \cite{HouseStevens1956, LindblomSundberg1971, LadefogedEtAl1978}, and F4 with high front vowel distinctions and speaker voice quality \cite{EekMeister1994}. 
Vowel duration can also signal vowel quality, and denotes lexical differences in many languages.\looseness=-1

%-----------------------------------------
%-----------------------------------------

We extracted formant and duration information from each vowel using Praat \cite{praat}. 
The first four formants (F1--F4) were measured at each quartile and decile of the vowel. 
Formant estimation was performed with the Burg algorithm in Praat with pre-emphasis from 50 Hz, a time window of 25 ms, a time step of 6.25 ms, a maximum of five formants permitted, and a formant ceiling of 5000 Hz, which is the recommended value for a male vocal tract \cite{praat}. Note that the speakers in this corpus are predominantly male.\looseness=-1

%-----------------------------------------
\subsubsection{Sibilant measures}
\label{sec:sibs}

Standard phonetic measurements of sibilant fricatives such as \phon{s}, \phon{z}, \phon{S}, and \phon{Z} include measures of spectral shape, and also segment duration. Measures of spectral shape frequently distinguish sibilant place of articulation: higher concentrations of energy generally reflect more anterior constriction locations (e.g., \phon{s z} are produced closer to the teeth than \phon{S Z}). 
Segment duration can also signal contrasts in voicing status \cite{JongmanEtAl2000}.\looseness=-1

Our release contains the segment duration, spectral peak, the spectral moments of the frequency distribution (center of gravity: COG, variance, skewness, and kurtosis), as well as two measures of the mid-frequency peak determined by sibilant quality.
These are the mid-frequency peak between 3000 and 7000 Hz for alveolar sibilants, and between 2000 and 6000 Hz for post-alveolar sibilants \cite{KoenigEtAl2013, Shadle2016}.
The spectral information was obtained via multitaper spectral analysis \cite{multitaper}, with a time-bandwidth parameter ($nw$) of 4 and 8 tapers ($k$) over the middle 50\% of the fricative \cite{Blacklock2004}. 
Measurements were made using the methods described in \citet{forrest1988statistical} for spectral moments and \citet{KoenigEtAl2013} for spectral peak varieties.\looseness=-1

\subsection{Computation times}

Generating phoneme-level alignments and extracting subsequent phonetic measures takes significant time, computational resources, and domain knowledge. 
Our release enables the community to use this data directly without these prerequisites. 
\cref{tab:compute} shows that the time to extract our resources, once methods have been developed, was more than 6 CPU years, primarily for training eHMM models.\looseness=-1 
%----------------
\begin{table}
\footnotesize
\begin{adjustbox}{width=\columnwidth}
\begin{tabular}{lrr}
\toprule
 &  \multicolumn{2}{c}{\bf  Computation Time}\\ % \normalsize
 \cmidrule(l){2-3}
\textbf{Resource} & \textit{Per Language} & \textit{Total Time} \\ 
Utterance Alignments & 30m         & 14d 13h \\% 30m \\
Phoneme Alignments     & 3d 3h 37m   & 6y 12d 16h \\ %12m \\ %2202d!
Vowel Measures       & 45m         & 21d 20h \\ %15m \\
Sibilant Measures   & 20m         & 9d 17h \\ %~~0m\\
 \cmidrule(l){2-2} \cmidrule(l){3-3}
 & 3d 5h 0m & \bf 6y 58d 19h \\
\bottomrule
\end{tabular}
\end{adjustbox}
\caption{Computation time to generate the full corpus.}
\label{tab:compute}
\end{table}
%----------

%-----------------------------------------
\subsection{General caveats} 
\label{sec:caveats}

We caution that our labeling and alignment of the corpus contains errors. 
In particular, 
it is difficult to responsibly draw firm linguistic conclusions from the Unitran-based segments (\cref{sec:unitran}). In \cref{sec:future} we suggest future work to address these issues.\looseness=-1

\begin{enumerate}[label=\textbf{\Alph*},leftmargin=12pt]
    \itemsep0em 
    % -- utterance alignments
    \item\label{caveat:phonelabels} \textbf{Quality of Utterance Pairs}:
    For some utterances, the speech does not correspond completely to the text, due to incorrect co-segmentation.  In our phonetic studies, we threshold using reading-level MCD as a heuristic for overall alignment quality, and further threshold remaining readings using utterance-level MCD. We recommend others do so as well. 
    %
    % -- phone alignment and phone label quality: orthography
    \item\label{caveat:phonelabels} \textbf{Phoneme Label Consistency and Accuracy}: 
    Phoneme-level labels are predicted from text without the aid of audio using G2P methods.  This may lead to systematic errors.    
    In particular, Unitran relies on a `universal' table that maps grapheme \graph{s} (for example) to phoneme \phon{s} in every context and every language.
    This is problematic for languages that use \graph{s} in some or all contexts to refer to other phonemes such as \phon{S} or \phon{\:s}, or use digraphs that contain \graph{s}, such as \graph{sh} for \phon{S}.  Thus, the predicted label \phon{s} may not consistently refer to the same phoneme within a language, nor to phonetically similar phonemes across languages. 
    Even WikiPron annotations are user-submitted and may not be
    internally consistent (e.g., some words use \phon{d Z} or \phon{t} while others use \phon{\textdyoghlig} or
    /\textsubbridge{\textipa{t}}/), nor comparable across languages.

`Phoneme' inventories for Unitran and WikiPron have been implicitly chosen by whoever designed the language's orthography or its WikiPron pages; while this may reflect a reasonable folk phonology, it may not correspond to the inventory of underlying or surface phonemes that any linguist would be likely to posit.\looseness=-1
    
    % -- MCD influenced by gender and language and TTS model quality
    \item\label{caveat:assessment} \textbf{Label and Alignment Assessment}: 
    While alignment quality for languages with Epitran and WikiPron can be assessed and calibrated beyond this corpus, it cannot for those languages with only Unitran alignments; the error rate on languages without resources to evaluate PER is unknown to us. 
    The Unitran alignments should be treated as a first-pass alignment which may still be useful for a researcher who is willing to perform quality control and correction of the alignments using automatic or manual procedures. 
    Our automatically-generated alignment offers an initial label and placement of the boundaries that would hopefully facilitate downstream analysis. 
    
    % -- num segment examples and utterances by lang; -- largely male and multi-speaker
    \item\label{caveat:representation} \textbf{Corpus Representation}: It is difficult to draw conclusions about `average behavior' across languages. Some language families are better represented in the corpus than others, with more languages, more Bible readings per language, more hours of speech per reading, or more examples of a given phoneme of interest.\footnote{See our \href{https://voxclamantisproject.github.io}{corpus website} for exact numbers of utterances and our phonetic measures per each language.} Additionally, the recordings by language are largely single-speaker (and predominantly male). This means that we can often draw conclusions only about a particular speaker's idiolect, rather than the population of speakers of the language.  Metadata giving the exact number of different speakers per recording do not exist. 
\end{enumerate}

%-----------------------------------------------------------------------------------------------------------
\section{Phonetic Case Studies}\label{sec:case-studies}
\label{sec:casestudies}
We present two case studies to illustrate the utility of our resource for exploration of cross-linguistic typology.
Phoneticians have posited several typological principles that may structure phonetic systems. 
Though previous research has provided some indication as to the direction and magnitude of expected effects, many instances of the principles have not yet been explored at scale. 
Our case studies investigate how well they account for cross-linguistic variation and systematicity for our phonetic measures from vowels and sibilants.
Below we present the data filtering methods for our case studies, followed by an introduction to and evaluation of phonetic dispersion and uniformity.\looseness=-1

%------------------------------------
\subsection{Data filtering}
\label{sec:data_filtering}
%------------------------------------
For quality, we use only the tokens extracted using high-resource pronunciations (Epitran and WikiPron) and only in languages with mean MCD lower than 8.0.\footnote{In the high-MCD languages, even the low-MCD utterances seem to be untrustworthy.} Furthermore, we only use those utterances with MCD lower than 6.0. 
The vowel analyses focus on F1 and F2 in ERB taken at the vowel midpoint \cite{ZwickerTerhardt1980, GlasbergMoore1990}.\footnote{The Equivalent Rectangular Bandwidth (ERB) scale is a psychoacoustic scale that better approximates human perception, which may serve as auditory feedback for the phonetic realization \cite{Fletcher1923, Nearey1978, ZwickerTerhardt1980, GlasbergMoore1990}. The precise equation comes from \citet[][Eq. 4]{GlasbergMoore1990}.} 
The sibilant analyses focus on mid-frequency peak of \phon{s} and \phon{z}, also in ERB. 
Vowel tokens with F1 or F2 measures beyond two standard deviations from the label- and reading-specific mean were excluded, as were tokens for which Praat failed to find a measurable F1 or F2, or whose duration exceeded 300 ms.
Sibilant tokens with mid-frequency peak or duration measures beyond two standard deviations from the label- and reading-specific mean were also excluded.
When comparing realizations of two labels such as \phon{i}--\phon{u} or \phon{s}--\phon{z}, we excluded readings that did not contain at least 50 tokens of each label.
We show data representation with different filtering methods in \cref{app:retention}.\looseness=-1

After filtering, the vowel analyses included 48 readings covering 38 languages and 11 language families. The distribution of language families was 21 Indo-European, 11 Austronesian, 3 Creole/Pidgin, 3 Turkic, 2 Afro-Asiatic, 2 Tai-Kadai, 2 Uto-Aztecan, 1 Austro-Asiatic, 1 Dravidian, 1 Hmong-Mien, and 1 Uralic. Approximately 8.2 million vowel tokens remained, with a minimum of $\approx$31,000 vowel tokens per reading. The sibilant analysis included 22 readings covering 18 languages and 6 language families. The distribution of language families was 10 Indo-European, 6 Austronesian, 3 Turkic, 1 Afro-Asiatic, 1 Austro-Asiatic, and 1 Creole/Pidgin. The decrease in total number of readings relative to the vowel analysis primarily reflects the infrequency of \phon{z} cross-linguistically.
Approximately 385,000 /s/ and 83,000 /z/ tokens remained, with a minimum of $\approx$5,200 tokens per reading.\looseness=-1

%------------------------------------
\subsection{Phonetic dispersion} 
\label{sec:dispersion}
Phonetic dispersion refers to the principle that contrasting speech sounds should be distinct from one another in phonetic space \cite{Martinet1955, Jakobson1968, Flemming1995, Flemming2004}.
Most studies investigating this principle have focused on its validity within vowel systems, as we do here.
While languages tend to have seemingly well-dispersed vowel inventories such as \{\phon{i}, \phon{a}, \phon{u}\} \cite{Joos1948, StevensKeyser2010}, the actual phonetic realization of each vowel can vary substantially \cite{LindauWood1977, Disner1983}. 
One prediction of dispersion is that the number of vowel categories in a language should be inversely related to the degree of per-category acoustic variation \cite{Lindblom1986}. Subsequent findings have cast doubt on this \cite{Livijn2000, RecasensEspinosa2009, VauxSamuels2015}, but these studies have been limited by the number and diversity of languages investigated.

To investigate this, we measured the correlation between the number of vowel categories in a language and the degree of per-category variation, as measured by the \emph{joint} entropy of (F1, F2) conditioned on the vowel category.
We model $p(\text{F1}, \text{F2} \mid V)$ using a bivariate Gaussian for each vowel type $v$.
We can then compute the joint conditional entropy under this model as $\ent (\text{F1}, \text{F2} \mid V)= \sum_v p(v)\,\ent (\text{F1}, \text{F2} \mid V=v) = \sum_v p(v) \frac{1}{2} \ln \det (2 \pi e \Sigma_v)$, where $\Sigma_v$ is the covariance matrix for the model of vowel $v$.

Vowel inventory sizes per reading ranged from 4 to 20 vowels, with a median of 8. Both Spearman and Pearson correlations between entropy estimate and vowel inventory size across analyzed languages were small and not significant (Spearman $\rho$ = 0.11, $p = 0.44$; Pearson $r$ = 0.11, $p = 0.46$), corroborating previous accounts of the relationship described in \citet{Livijn2000} and \citet{VauxSamuels2015} with a larger number of languages---a larger vowel inventory does not necessarily imply more precision in vowel category production.\footnote{Since differential entropy is sensitive to parameterization, we also measured this correlation using formants in hertz, instead of in ERB, as ERB is on a logarithmic scale. This change did not the influence the pattern of results (Spearman $\rho$ = 0.12, $p = 0.41$; Pearson $r$ = 0.13, $p = 0.39$).}

%-----------------------------------------
\begin{figure*}[!th]
\centering
\begin{subfigure}[b]{.39\textwidth}
    \centering
    \includegraphics[width=\textwidth]{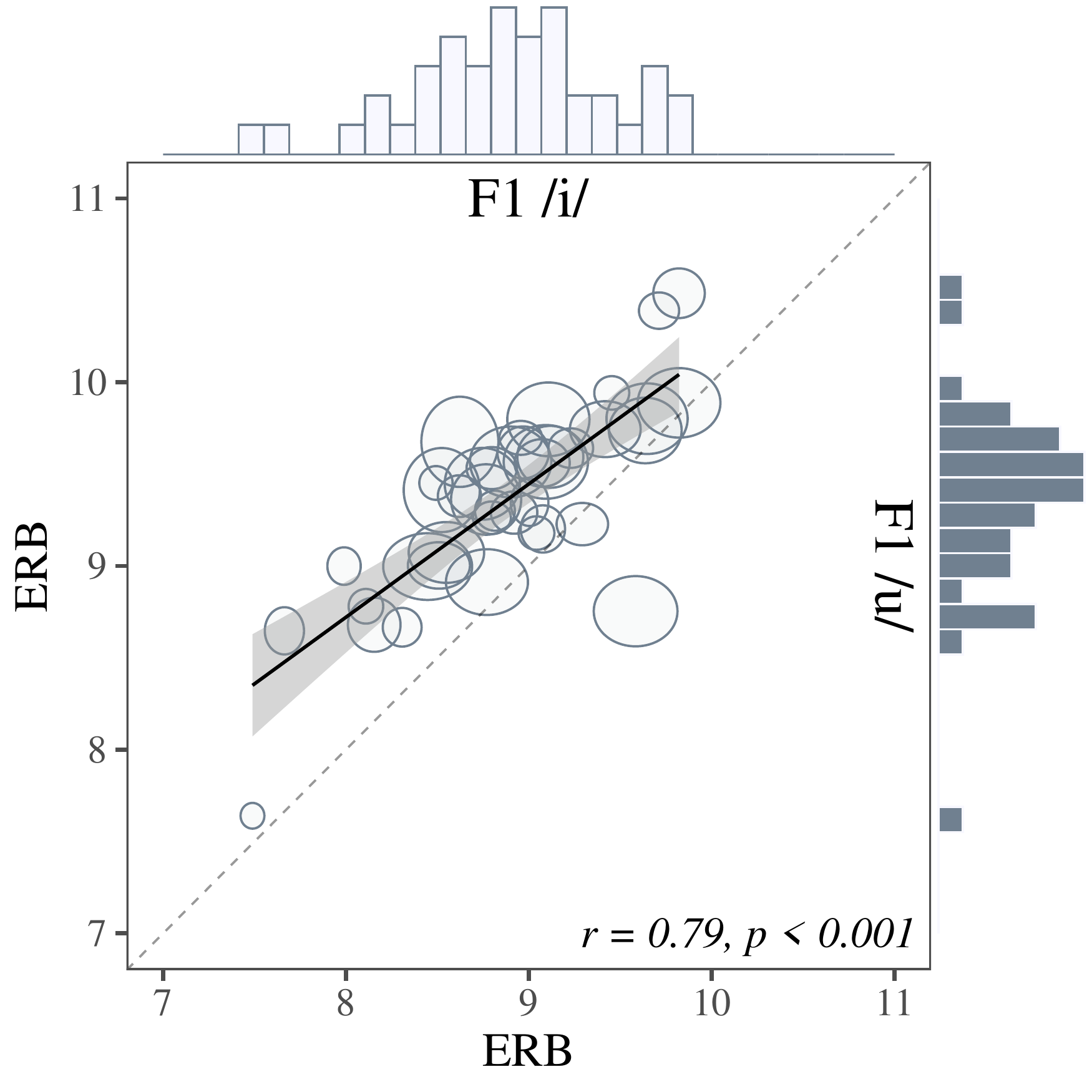}%
    \caption{\textbf{F1 of \phon{i}--\phon{u} in ERB}\label{fig:iu}}
\end{subfigure}
\qquad\qquad
\centering
\begin{subfigure}[b]{.39\textwidth}
    \centering
    \includegraphics[width=\textwidth]{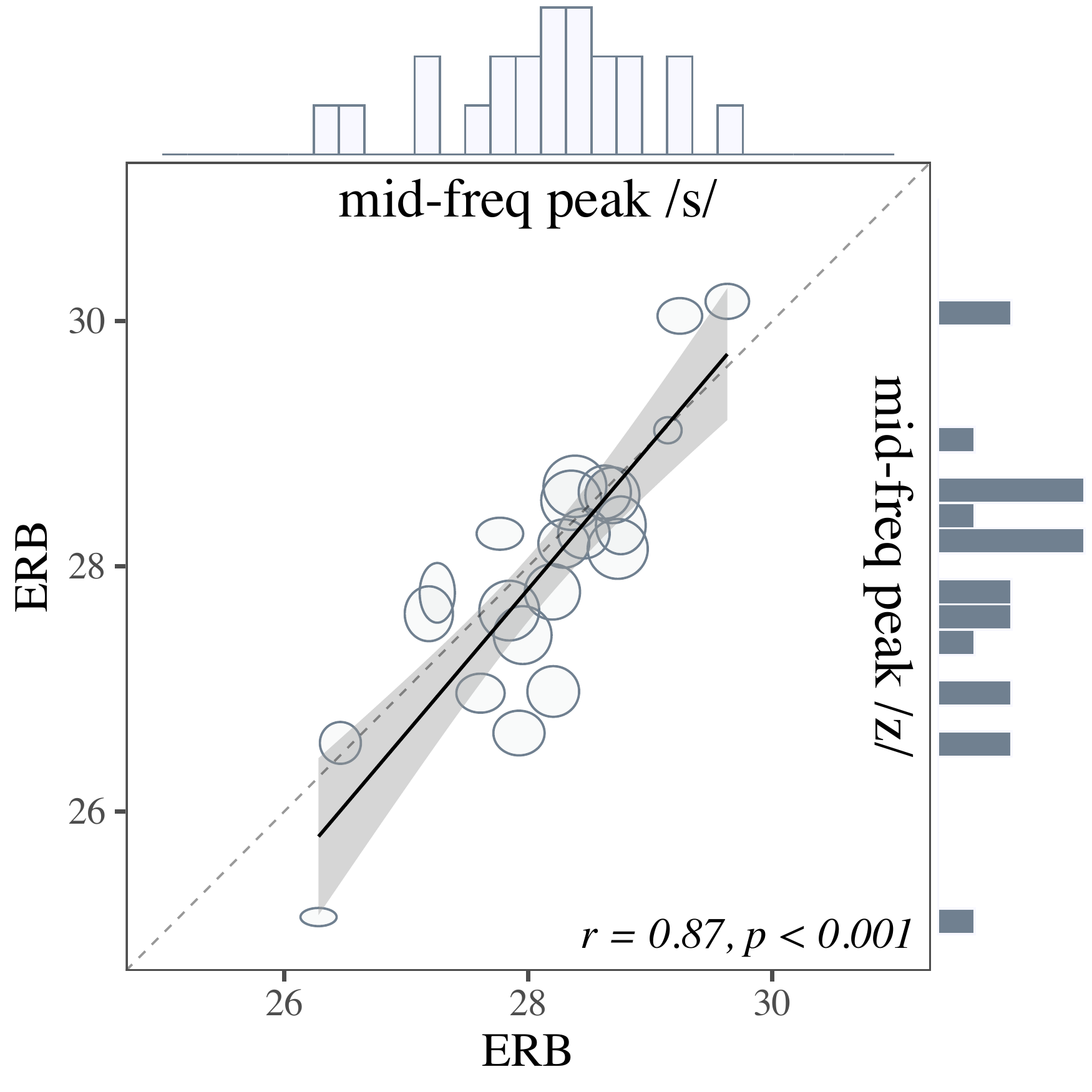}%
    \caption{\textbf{Mid-frequency peak of \phon{s}--\phon{z} in ERB}}\label{fig:uo}
\end{subfigure} 
\caption{Correlations of mean F1 (ERB) between /i/ and /u/ and of mean mid-frequency peak (ERB) between \phon{s} and \phon{z}. The paired segments share a relevant phonological feature specification that is approximated by the acoustic-phonetic measurement: vowel height by F1 and sibilant place by mid-frequency peak. Each reading is represented by an ellipsoid, centered on the paired means and shaped by $\frac{1}{10}$ of their respective standard deviations. The solid line reflects the best-fit linear regression line with standard error in gray shading; the dashed line shows the line of equality. Marginal histograms show the range of variation in the segment-specific means.}
\label{fig:doublefig}
\end{figure*}
%-----------------------------------------

%-----------------------------------------
\subsection{Phonetic uniformity} 
\label{sec:uniformity}
Previous work suggests that F1 is fairly uniform with respect to phonological height. 
Within a single language,  the mean F1s of \phon{e} and \phon{o}---which share a height---have been found to be correlated across speakers (\textit{Yorkshire English}: \citealp{Watt2000}; \textit{French}: \citealp{MenardEtAl2008}; \textit{Brazilian Portuguese}: \citealp{Oushiro2019}; \textit{Dutch, English, French, Japanese, Portuguese, Spanish}: \citealp{SchwartzMenard2019}). 
Though it is physically possible for these vowels to differ in F1 realization, the correlations indicate a strong tendency for languages and individual speakers to yoke these two representations together.\looseness=-1

Systematicity in the realization of sibilant place of articulation has also been observed across speakers of American English and Czech \cite{Chodroff2017}. Phonetic correlates of sibilant place strongly covary between \phon{s} and \phon{z}, which share a [+anterior] place of articulation and are produced the alveolar ridge, and between \phon{S} and \phon{Z}, which share a [-anterior] place of articulation and are produced behind the alveolar ridge.\looseness=-1 

A principle of uniformity may account for these above findings. Uniformity here refers to a principle in which a distinctive phonological feature should have a consistent phonetic realization, within a language or speaker, across different segments with that feature \cite{Keating2003, ChodroffEtAl2019}. Similar principles posited in the literature include Maximal Use of Available Controls, in which a control refers to an integrated perceptual and motor phonetic target \cite{MenardEtAl2008}, as well as a principle of gestural economy \cite{Maddieson1995}. Phonetic realization refers to the mapping from the abstract distinctive feature to an abstract phonetic target. We approximate this phonetic target via an acoustic-phonetic measurement, but we emphasize that the acoustic measurement is not necessarily a direct reflection of an underlying phonetic target (which could be an articulatory gesture, auditory goal, or perceptuo-motor representation of the sound). We make the simplifying assumption that the acoustic-phonetic formants (F1, F2) directly correspond to phonetic targets linked to the vowel features of height and backness.\looseness=-1 

More precisely, uniformity of a phonetic measure with respect to a phonological feature means that any two segments sharing that feature will tend to have approximately equal measurements in a given language, even when that value varies across languages.  We can observe whether this is true by plotting the measures of the two segments against each other by language (e.g., \cref{fig:doublefig}).

%-----------------------------------------
\paragraph{Vowels.} 

As shown in \cref{fig:doublefig} and \cref{tab:f1-cors}, the strongest correlations in mean F1 frequently reflected uniformity of height (e.g., high vowels \phon{i}--\phon{u}: $r$ = 0.79,  $p < 0.001$, mid vowels \phon{e}--\phon{o}: $r$ = 0.62, $p < 0.01$).\footnote{$p$-values are corrected for multiple comparisons using the Benjamini-Hochberg correction and a false discovery rate of 0.25 \cite{benjamini1995controlling}.}
Nevertheless, some vowel pairs that differed in height were also moderately correlated in mean F1 (e.g., \phon{o}--\phon{a}: $r = 0.66$, $p < 0.001$).
Correlations of mean F1 were overall moderate in strength, regardless of the vowels' phonological specifications. 

Correlations of mean F2 were also strongest among vowels with a uniform backness specification (e.g., back vowels \phon{u}--\phon{o}: $r = 0.69$, $p < 0.001$; front vowels \phon{i}--\phon{E}: $r = 0.69$, $p < 0.05$; \cref{tab:f2-cors}). 
The correlation between front tense vowels \phon{i} and \phon{e} was significant and in the expected direction, but also slightly weaker than the homologous back vowel pair ($r = 0.41$, $p < 0.05$). 
Vowels differing in backness frequently had negative correlations, which could reflect influences of category crowding or language-/speaker-specific differences in peripheralization.
We leave further exploration of those relationships to future study.\looseness=-1

The moderate to strong F1 correlations among vowels with a shared height specification are consistent with expectations based on previous studies, and also with predictions of uniformity. 
Similarly, we find an expected correlation of F2 means for vowels with a shared height specification.
The correlations of vowel pairs that were predicted to have significant correlations, but did not, tended to have small sample sizes ($< 14$ readings).

Nevertheless, the correlations are not perfect; nor are the patterns. 
For instance, the back vowel correlations of F2 are stronger than the front vowel correlations.
While speculative, the apparent peripheralization of \phon{i} (as revealed in the negative F2 correlations) could have weakened the expected uniformity relation of \phon{i} with other front vowels.
Future research should take into account additional influences of the vowel inventory composition, as well as articulatory or auditory factors for a more complete understanding of the structural forces in the phonetic realization of vowels.\looseness=-1

%-----------------------------------------

\paragraph{Sibilants.}
The mean mid-frequency peak values for \phon{s} and \phon{z} each varied substantially across readings, and were also strongly correlated with one another ($r = 0.87$, $p < 0.001$; \cref{fig:doublefig}).\footnote{The magnitude of this correlation did not change when using hertz ($r = 0.86$, $p < 0.001$).} 
This finding suggests a further influence of uniformity on the realization of place for /s/ and /z/, and the magnitude is comparable to previous correlations observed across American English and Czech speakers, in which $r$ was $\approx$0.90 \cite{Chodroff2017}.

%-----------------------------------------
\section{Directions for Future Work}
\label{sec:future}

We hope our corpus may serve as a touchstone for further improvements in phonetic typology research and methodology. 
Here we suggest potential steps forward for known areas (\cref{sec:caveats}) where this corpus could be improved: 

\begin{enumerate}[label=\textbf{\Alph*},leftmargin=12pt]
    \itemsep0em 
    % -- `sentences' -- 
    \item \textbf{Sentence alignments} were generated using Unitran, and could be improved with higher-quality G2P and verse-level text segmentation to standardize utterances across languages.\looseness=-1
    %
    % -- consistency and accuracy of phoneme-level labels -- 
    \item \textbf{Consistent and comparable phoneme labels} are the ultimate goal.  Concurrent work on universal phone recognition \citep{li2020universal} addresses this issue through a universal phone inventory constrained by language-specific PHOIBLE inventories \citep{phoible}. However, free-decoding phones from speech alone is challenging.
    One exciting possibility is to use the orthography and audio jointly to guide semi-supervised learning of per-language pronunciation lexicons \cite{lu2013acoustic, zhang2017acoustic}.\looseness=-1
    %
    % -- quality assessment -- 
    \item \textbf{Reliable quality assessment} for current methods remains an outstanding research question for many languages. 
    For covered languages, using a universal label set to map additional high quality lexicons (e.g., hand-annotated lexicons) to the same label space as ours would enable direct label and alignment assessment through precision, recall, and PER.\looseness=-1  
    \item \textbf{Curating additional resources} beyond this corpus would improve coverage and balance, such as contributing additional Epitran modules. 
    Additional readings exist for many languages on the original \url{bible.is} site and elsewhere. 
    Annotations with speaker information are not available, but improved unsupervised speaker clustering may also support better analysis.\looseness=-1 
\end{enumerate}

%-----------------------------------------
\section{Conclusion}

\corpusname is the first large-scale corpus for phonetic typology, with extracted phonetic features for \numisos typologically diverse languages. 
We present two case studies illustrating both the research potential and limitations of this corpus for investigation of phonetic typology at a large scale. 
We discuss several caveats for the use of this corpus and areas for substantial improvement.
Nonetheless, we hope that directly releasing our alignments and token-level features enables greater research accessibility in this area. 
We hope this corpus will motivate and enable further developments in both phonetic typology and methodology for working with cross-linguistic speech corpora.\looseness=-1

%---------------------------
\section*{Acknowledgments}
The authors gratefully acknowledge Colin Wilson for his guidance and discussion on the topic, Florian Metze for resources, and Carlos Aguirre for helpful feedback.

\bibliography{acl2020,bibliography}

\begin{thebibliography}{65}
\expandafter\ifx\csname natexlab\endcsname\relax\def\natexlab#1{#1}\fi

\bibitem[{Anumanchipalli et~al.(2011)Anumanchipalli, Prahallad, and
  Black}]{anumanchipalli2011festvox}
Gopala~Krishna Anumanchipalli, Kishore Prahallad, and Alan~W. Black. 2011.
\newblock \href {https://www.cs.cmu.edu/~awb/papers/vlsp2011_festvox.pdf}
  {Festvox: Tools for creation and analyses of large speech corpora}.
\newblock In \emph{Workshop on Very Large Scale Phonetics Research, UPenn,
  Philadelphia}.

\bibitem[{Ardila et~al.(2020)Ardila, Branson, Davis, Henretty, Kohler, Meyer,
  Morais, Saunders, Tyers, and Weber}]{ardila2019common}
Rosana Ardila, Megan Branson, Kelly Davis, Michael Henretty, Michael Kohler,
  Josh Meyer, Reuben Morais, Lindsay Saunders, Francis~M. Tyers, and Gregor
  Weber. 2020.
\newblock \href {http://arxiv.org/abs/1912.06670} {Common {V}oice: A
  massively-multilingual speech corpus}.
\newblock In \emph{Proceedings of the Twelfth International Conference on
  Language Resources and Evaluation (LREC 2020)}.

\bibitem[{Becker-Kristal(2010)}]{BeckerKristal2010}
Roy Becker-Kristal. 2010.
\newblock \href {https://search.proquest.com/docview/861321787} {\emph{Acoustic
  typology of vowel inventories and Dispersion Theory: Insights from a large
  cross-linguistic corpus}}.
\newblock Ph.D. thesis, University of California, Los Angeles.

\bibitem[{Benjamini and Hochberg(1995)}]{benjamini1995controlling}
Yoav Benjamini and Yosef Hochberg. 1995.
\newblock \href {https://www.jstor.org/stable/2346101} {Controlling the false
  discovery rate: {A} practical and powerful approach to multiple testing}.
\newblock \emph{Journal of the Royal Statistical Society: Series B
  (Methodological)}, 57(1):289--300.

\bibitem[{Black(2006)}]{black2006clustergen}
Alan~W. Black. 2006.
\newblock {CLUSTERGEN}: A statistical parametric synthesizer using trajectory
  modeling.
\newblock In \emph{Proceedings of INTERSPEECH}.

\bibitem[{Black(2019)}]{black2019cmu}
Alan~W. Black. 2019.
\newblock \href {https://doi.org/10.1109/ICASSP.2019.8683536} {{CMU}
  {W}ilderness {M}ultilingual {S}peech {D}ataset}.
\newblock In \emph{ICASSP 2019-2019 IEEE International Conference on Acoustics,
  Speech and Signal Processing (ICASSP)}, pages 5971--5975, Brighton, UK. IEEE.

\bibitem[{Blacklock(2004)}]{Blacklock2004}
Oliver Blacklock. 2004.
\newblock \href {https://eprints.soton.ac.uk/420069/} {\emph{Characteristics of
  Variation in Production of Normal and Disordered Fricatives, Using
  Reduced-Variance Spectral Methods}}.
\newblock Ph.D. thesis, University of Southampton.

\bibitem[{Boersma and Weenink(2019)}]{praat}
Paul Boersma and David Weenink. 2019.
\newblock \href {http://www.fon.hum.uva.nl/praat/} {Praat: Doing phonetics by
  computer [computer program]. version 6.0.45}.

\bibitem[{Cho and Ladefoged(1999)}]{ChoLadefoged1999}
Taehong Cho and Peter Ladefoged. 1999.
\newblock \href {https://doi.org/10.1006/jpho.1999.0094} {Variation and
  universals in {VOT}: Evidence from 18 languages}.
\newblock \emph{Journal of Phonetics}, 27(2):207--229.

\bibitem[{Chodroff(2017)}]{Chodroff2017}
Eleanor Chodroff. 2017.
\newblock \href
  {https://www.eleanorchodroff.com/articles/Chodroff-Dissertation-2017_Final.pdf}
  {\emph{Structured Variation in Obstruent Production and Perception}}.
\newblock Ph.D. thesis, Johns Hopkins University.

\bibitem[{Chodroff et~al.(2019)Chodroff, Golden, and Wilson}]{ChodroffEtAl2019}
Eleanor Chodroff, Alessandra Golden, and Colin Wilson. 2019.
\newblock \href {https://asa.scitation.org/doi/10.1121/1.5088035} {Covariation
  of stop voice onset time across languages: {E}vidence for a universal
  constraint on phonetic realization}.
\newblock \emph{The Journal of the Acoustical Society of America},
  145(1):EL109--EL115.

\bibitem[{Disner(1983)}]{Disner1983}
Sandra~Ferrari Disner. 1983.
\newblock \emph{Vowel Quality: {T}he Relation between Universal and
  Language-specific Factors}.
\newblock Ph.D. thesis, UCLA.

\bibitem[{Eberhard and Fennig(2020)}]{ethnologue2020}
Gary F.~Simons Eberhard, David~M. and Charles~D. Fennig, editors. 2020.
\newblock \href {http://www.ethnologue.com} {\emph{Ethnologue: Languages of the
  world}}, 23 edition.
\newblock SIL international.
\newblock Online version: http://www.ethnologue.com.

\bibitem[{Eek and Meister(1994)}]{EekMeister1994}
Arvo Eek and Einar Meister. 1994.
\newblock \href
  {https://www.researchgate.net/profile/Eugene_Buder/publication/263965982_Cross-Language_Differences_in_Phonological_Acquisition_Swedish_and_American_t/links/568c7a6908ae153299b668f3.pdf#page=65}
  {Acoustics and perception of {E}stonian vowel types}.
\newblock \emph{Phonetic Experimental Research}, XVIII:146--158.

\bibitem[{Engstrand and Cunningham-Andersson(1988)}]{Engstrand1988}
Olle Engstrand and Una Cunningham-Andersson. 1988.
\newblock \href {https://ir.canterbury.ac.nz/handle/10092/11791} {Iris - a data
  base for cross-linguistic phonetic research}.

\bibitem[{Flemming(1995)}]{Flemming1995}
Edward~S. Flemming. 1995.
\newblock \href
  {https://linguistics.ucla.edu/general/dissertations/Flemming.1995.pdf}
  {\emph{Auditory Representations in Phonology}}.
\newblock Ph.D. thesis, UCLA.

\bibitem[{Flemming(2004)}]{Flemming2004}
Edward~S. Flemming. 2004.
\newblock \href {https://doi.org/10.1017/CBO9780511486401.008} {Contrast and
  perceptual distinctiveness}.
\newblock In Bruce Hayes, R.~Kirchner, and Donca Steriade, editors, \emph{The
  Phonetic Bases of Phonological Markedness}, 1968, pages 232--276. University
  Press, Cambridge, MA.

\bibitem[{Fletcher(1923)}]{Fletcher1923}
Harvey Fletcher. 1923.
\newblock \href {https://doi.org/https://doi.org/10.1016/S0016-0032(23)91006-9}
  {Physical measurements of audition and their bearing on the theory of
  hearing}.
\newblock \emph{Journal of the Franklin Institute}, 196(3):289--326.

\bibitem[{Forrest et~al.(1988)Forrest, Weismer, Milenkovic, and
  Dougall}]{forrest1988statistical}
Karen Forrest, Gary Weismer, Paul Milenkovic, and Ronald~N. Dougall. 1988.
\newblock \href {https://doi.org/10.1121/1.396977} {Statistical analysis of
  word-initial voiceless obstruents: Preliminary data}.
\newblock \emph{The Journal of the Acoustical Society of America},
  84(1):115--123.

\bibitem[{Glasberg and Moore(1990)}]{GlasbergMoore1990}
Brian~R. Glasberg and Brian~C.J. Moore. 1990.
\newblock \href {https://doi.org/10.1016/0378-5955(90)90170-T} {{Derivation of
  auditory filter shapes from notched-noise data}}.
\newblock \emph{Hearing Research}, 47(1-2):103--138.

\bibitem[{Gordon and Roettger(2017)}]{GordonRoettger2017}
Matthew Gordon and Timo Roettger. 2017.
\newblock \href {https://doi.org/https://doi.org/10.1515/lingvan-2017-0007}
  {Acoustic correlates of word stress: A cross-linguistic survey}.
\newblock \emph{Linguistics Vanguard}, 3(1).

\bibitem[{Gorman et~al.(2020)Gorman, Ashby, Goyzueta, McCarthy, Wu, and
  You}]{gorman2020sigmorphon}
Kyle Gorman, Lucas~F.E. Ashby, Aaron Goyzueta, Arya~D. McCarthy, Shijie Wu, and
  Daniel You. 2020.
\newblock The {SIGMORPHON} 2020 shared task on multilingual grapheme-to-phoneme
  conversion.
\newblock In \emph{Proceedings of the {SIGMORPHON} Workshop}.

\bibitem[{Harper(2011)}]{babel2011iarpa}
Mary Harper. 2011.
\newblock \href {https://www.iarpa.gov/index.php/research-programs/babel} {The
  {IARPA} {B}abel multilingual speech database}.
\newblock Accessed: 2020-05-01.

\bibitem[{House and Stevens(1956)}]{HouseStevens1956}
Arthur~S. House and Kenneth~N. Stevens. 1956.
\newblock \href {https://doi.org/10.1044/jshd.2102.218} {Analog studies of the
  nasalization of vowels}.
\newblock \emph{The Journal of Speech and Hearing Disorders}, 21(2):218--232.

\bibitem[{Jakobson(1968)}]{Jakobson1968}
Roman Jakobson. 1968.
\newblock \emph{Child Language, Aphasia and Phonological Universals}.
\newblock Mouton Publishers.

\bibitem[{Jongman et~al.(2000)Jongman, Wayland, and Wong}]{JongmanEtAl2000}
Allard Jongman, Ratree Wayland, and Serena Wong. 2000.
\newblock \href {https://doi.org/10.1121/1.1288413} {{Acoustic characteristics
  of {E}nglish fricatives}}.
\newblock \emph{The Journal of the Acoustical Society of America},
  108(3):1252--1263.

\bibitem[{Joos(1948)}]{Joos1948}
Martin Joos. 1948.
\newblock \href {https://doi.org/10.2307/522229} {Acoustic phonetics}.
\newblock \emph{Language}, 24(2):5--136.

\bibitem[{Keating(2003)}]{Keating2003}
Patricia~A. Keating. 2003.
\newblock \href
  {https://www.internationalphoneticassociation.org/icphs-proceedings/ICPhS2003/papers/p15_0375.pdf}
  {Phonetic and other influences on voicing contrasts}.
\newblock In \emph{Proceedings of the 15th International Congress of Phonetic
  Sciences}, pages 20--23, Barcelona, Spain.

\bibitem[{Koenig et~al.(2013)Koenig, Shadle, Preston, and
  Mooshammer}]{KoenigEtAl2013}
Laura Koenig, Christine~H. Shadle, Jonathan~L. Preston, and Christine~R.
  Mooshammer. 2013.
\newblock \href {https://doi.org/10.1044/1092-4388(2012/12-0038)} {Toward
  improved spectral measures of /s/: {R}esults from adolescents}.
\newblock \emph{Journal of Speech, Language, and Hearing Research},
  56(4):1175--1189.

\bibitem[{Kominek et~al.(2008)Kominek, Schultz, and
  Black}]{kominek2008synthesizer}
John Kominek, Tanja Schultz, and Alan~W. Black. 2008.
\newblock \href
  {https://www.semanticscholar.org/paper/Synthesizer-voice-quality-of-new-languages-with-mel-Kominek-Schultz/04ff7bb2bc3bea9150bc6f2902adac3cd54b26b8}
  {Synthesizer voice quality of new languages calibrated with mean mel cepstral
  distortion}.
\newblock In \emph{Spoken Languages Technologies for Under-Resourced
  Languages}.

\bibitem[{Ladefoged et~al.(1978)Ladefoged, Harshman, Goldstein, and
  Rice}]{LadefogedEtAl1978}
Peter Ladefoged, Richard Harshman, Louis Goldstein, and Lloyd Rice. 1978.
\newblock \href {https://doi.org/10.1121/1.382086} {{Generating vocal tract
  shapes from formant frequencies}}.
\newblock \emph{The Journal of the Acoustical Society of America},
  64(4):1027--1035.

\bibitem[{Ladefoged and Johnson(2014)}]{LadefogedJohnson2014}
Peter Ladefoged and Keith Johnson. 2014.
\newblock \emph{A Course in Phonetics}.
\newblock Nelson Education.

\bibitem[{Ladefoged and Maddieson(2007)}]{UCLAPhoneticsLab}
Peter Ladefoged and Ian Maddieson. 2007.
\newblock \href {http://archive.phonetics.ucla.edu/} {The {UCLA} phonetics lab
  archive}.

\bibitem[{Lee et~al.(2020)Lee, Ashby, Garza, Lee-Sikka, Miller, Wong, McCarthy,
  and Gorman}]{wikipron}
Jackson~L. Lee, Lucas~F.E. Ashby, M.~Elizabeth Garza, Yeonju Lee-Sikka, Sean
  Miller, Alan Wong, Arya~D. McCarthy, and Kyle Gorman. 2020.
\newblock Massively multilingual pronunciation mining with {W}iki{P}ron.
\newblock In \emph{Proceedings of the Twelfth International Conference on
  Language Resources and Evaluation (LREC 2020)}. European Language Resources
  Association (ELRA).
\newblock Resources downloadable from
  \url{https://github.com/kylebgorman/wikipron}.

\bibitem[{Li et~al.(2020)Li, Dalmia, Li, Lee, Littell, Yao, Anastasopoulos,
  Mortensen, Neubig, Black et~al.}]{li2020universal}
Xinjian Li, Siddharth Dalmia, Juncheng Li, Matthew Lee, Patrick Littell, Jiali
  Yao, Antonios Anastasopoulos, David~R Mortensen, Graham Neubig, Alan~W.
  Black, et~al. 2020.
\newblock \href {https://ieeexplore.ieee.org/document/9054362} {Universal phone
  recognition with a multilingual allophone system}.
\newblock In \emph{Proceedings of IEEE International Conference on Acoustics,
  Speech and Signal Processing (ICASSP)}, pages 8249--8253. IEEE.

\bibitem[{Lindau and Wood(1977)}]{LindauWood1977}
Mona Lindau and Patricia Wood. 1977.
\newblock \href {https://escholarship.org/content/qt8nj8480j/qt8nj8480j.pdf}
  {Acoustic vowel spaces}.
\newblock \emph{UCLA Working Papers in Phonetics}, 38:41--48.

\bibitem[{Lindblom(1986)}]{Lindblom1986}
Bj{\"{o}}rn Lindblom. 1986.
\newblock {Phonetic universals in vowel systems}.
\newblock In John~J. Ohala and Jeri Jaeger, editors, \emph{Experimental
  Phonology}, pages 13--44. Academic Press, Orlando.

\bibitem[{Lindblom and Sundberg(1971)}]{LindblomSundberg1971}
Bj{\"{o}}rn Lindblom and Johan Sundberg. 1971.
\newblock \href {https://doi.org/10.1121/1.1912750} {Acoustical consequences of
  lip, tongue, jaw, and larynx movement}.
\newblock \emph{The Journal of the Acoustical Society of America},
  50(4B):1166--1179.

\bibitem[{Livijn(2000)}]{Livijn2000}
Peder Livijn. 2000.
\newblock \href
  {http://citeseerx.ist.psu.edu/viewdoc/download?doi=10.1.1.28.3871&rep=rep1&type=pdf}
  {Acoustic distribution of vowels in differently sized inventories--hot spots
  or adaptive dispersion}.
\newblock \emph{Phonetic Experimental Research, Institute of Linguistics,
  University of Stockholm (PERILUS)}, 11.

\bibitem[{Lu et~al.(2013)Lu, Ghoshal, and Renals}]{lu2013acoustic}
Liang Lu, Arnab Ghoshal, and Steve Renals. 2013.
\newblock \href {https://doi.org/10.1109/ASRU.2013.6707759} {Acoustic
  data-driven pronunciation lexicon for large vocabulary speech recognition}.
\newblock In \emph{2013 IEEE Workshop on Automatic Speech Recognition and
  Understanding}, pages 374--379. IEEE.

\bibitem[{Maddieson(1995)}]{Maddieson1995}
Ian Maddieson. 1995.
\newblock Gestural economy.
\newblock In \emph{Proceedings of the 13th International Congress of Phonetic
  Sciences}, Stockholm, Sweden.

\bibitem[{Martinet(1955)}]{Martinet1955}
Andr{\'{e}} Martinet. 1955.
\newblock \emph{\'Economie Des Changements Phon\'etiques: Trait\'{e} de
  Phonologie Diachronique}, volume~10.
\newblock Bibliotheca Romanica.

\bibitem[{M{\'{e}}nard et~al.(2008)M{\'{e}}nard, Schwartz, and
  Aubin}]{MenardEtAl2008}
Lucie M{\'{e}}nard, Jean-Luc Schwartz, and J{\'{e}}r{\^{o}}me Aubin. 2008.
\newblock \href {https://doi.org/10.1016/j.specom.2007.06.004} {Invariance and
  variability in the production of the height feature in {F}rench vowels}.
\newblock \emph{Speech Communication}, 50:14--28.

\bibitem[{Moran and McCloy(2019)}]{phoible}
Steven Moran and Daniel McCloy, editors. 2019.
\newblock \href {https://phoible.org/} {\emph{PHOIBLE 2.0}}.
\newblock Max Planck Institute for the Science of Human History, Jena.

\bibitem[{Mortensen et~al.(2018)Mortensen, Dalmia, and
  Littell}]{epitran2018mortensen}
David~R. Mortensen, Siddharth Dalmia, and Patrick Littell. 2018.
\newblock \href {http://www.lrec-conf.org/proceedings/lrec2018/pdf/890.pdf}
  {Epitran: Precision {G2P} for many languages}.
\newblock In \emph{Proceedings of the Eleventh International Conference on
  Language Resources and Evaluation (LREC 2018)}, Paris, France. European
  Language Resources Association (ELRA).

\bibitem[{Nearey(1977)}]{Nearey1978}
Terrance~M. Nearey. 1977.
\newblock \emph{Phonetic Feature Systems for Vowels}.
\newblock Ph.D. thesis, University of Alberta.
\newblock Reprinted 1978 by Indiana University Linguistics Club.

\bibitem[{Novak et~al.(2016)Novak, Minematsu, and
  Hirose}]{novak2016phonetisaurus}
Josef~Robert Novak, Nobuaki Minematsu, and Keikichi Hirose. 2016.
\newblock \href {https://dx.doi.org/10.1017/S1351324915000315} {Phonetisaurus:
  Exploring grapheme-to-phoneme conversion with joint n-gram models in the
  {WFST} framework}.
\newblock \emph{Natural Language Engineering}, 22(6):907--938.

\bibitem[{Oushiro(2019)}]{Oushiro2019}
Livia Oushiro. 2019.
\newblock \href
  {http://www.assta.org/proceedings/ICPhS2019/papers/ICPhS_735.pdf} {Linguistic
  uniformity in the speech of {B}razilian internal migrants in a dialect
  contact situation}.
\newblock In \emph{Proceedings of the 19th International Congress of Phonetic
  Sciences, Melbourne, Australia 2019}, pages 686--690, Melbourne, Australia.
  Canberra, Australia: Australasian Speech Science and Technology Association
  Inc.

\bibitem[{Povey et~al.(2011)Povey, Ghoshal, Boulianne, Burget, Glembek, Goel,
  Hannemann, Motlicek, Qian, Schwarz, Silovsky, Stemmer, and
  Vesely}]{povey2011kaldi}
Daniel Povey, Arnab Ghoshal, Gilles Boulianne, Lukas Burget, Ondrej Glembek,
  Nagendra Goel, Mirko Hannemann, Petr Motlicek, Yanmin Qian, Petr Schwarz, Jan
  Silovsky, Georg Stemmer, and Karel Vesely. 2011.
\newblock \href
  {https://publications.idiap.ch/downloads/papers/2012/Povey_ASRU2011_2011.pdf}
  {The {K}aldi speech recognition toolkit}.
\newblock In \emph{IEEE 2011 Workshop on Automatic Speech Recognition and
  Understanding}. IEEE Signal Processing Society.
\newblock IEEE Catalog No.: CFP11SRW-USB.

\bibitem[{Prahallad et~al.(2006)Prahallad, Black, and Mosur}]{prahallad2006sub}
Kishore Prahallad, Alan~W. Black, and Ravishankhar Mosur. 2006.
\newblock \href {https://ieeexplore.ieee.org/document/1660155/} {Sub-phonetic
  modeling for capturing pronunciation variations for conversational speech
  synthesis}.
\newblock In \emph{Proceedings of IEEE International Conference on Acoustics,
  Speech and Signal Processing (ICASSP)}, volume~1. IEEE.

\bibitem[{Qian et~al.(2010)Qian, Hollingshead, Yoon, Kim, and
  Sproat}]{qian2010python}
Ting Qian, Kristy Hollingshead, Su-youn Yoon, Kyoung-young Kim, and Richard
  Sproat. 2010.
\newblock \href
  {http://www.lrec-conf.org/proceedings/lrec2010/pdf/30_Paper.pdf} {A {P}ython
  toolkit for universal transliteration}.
\newblock In \emph{Proceedings of the Seventh International Conference on
  Language Resources and Evaluation (LREC'10)}, Valletta, Malta. European
  Language Resources Association (ELRA).

\bibitem[{Rahim and Burr(2017)}]{multitaper}
Karim Rahim and Wesley~S. Burr. 2017.
\newblock \href
  {https://cran.r-project.org/web/packages/multitaper/multitaper.pdf}
  {{multitaper}: {M}ultitaper spectral analysis}.
\newblock \emph{R package version 1.0-14}.

\bibitem[{Recasens and Espinosa(2009)}]{RecasensEspinosa2009}
Daniel Recasens and Aina Espinosa. 2009.
\newblock \href {https://doi.org/10.1016/j.specom.2008.09.002} {Dispersion and
  variability in {C}atalan five and six peripheral vowel systems}.
\newblock \emph{Speech Communication}, 51(3):240--258.

\bibitem[{Schultz(2002)}]{schultz2002globalphone}
Tanja Schultz. 2002.
\newblock \href
  {https://www.isca-speech.org/archive/archive_papers/icslp_2002/i02_0345.pdf}
  {{G}lobal{P}hone: A multilingual speech and text database developed at
  {K}arlsruhe {U}niversity}.
\newblock In \emph{Seventh International Conference on Spoken Language
  Processing}, pages 345--348, Denver, CO.

\bibitem[{Schwartz and M{\'{e}}nard(2019)}]{SchwartzMenard2019}
Jean-Luc Schwartz and Lucie M{\'{e}}nard. 2019.
\newblock \href {https://doi.org/https://doi.org/10.31219/osf.io/b6rdv}
  {Structured idiosyncrasies in vowel systems}.
\newblock \emph{OSF Preprints}.

\bibitem[{Shadle et~al.(2016)Shadle, Chen, and Whalen}]{Shadle2016}
Christine~H. Shadle, Wei-rong Chen, and D.~H. Whalen. 2016.
\newblock \href {https://doi.org/10.1121/1.4970153} {{Stability of the main
  resonance frequency of fricatives despite changes in the first spectral
  moment}}.
\newblock \emph{The Journal of the Acoustical Society of America},
  140(4):3219--3220.

\bibitem[{Stevens and Keyser(2010)}]{StevensKeyser2010}
Kenneth~N. Stevens and Samuel~J. Keyser. 2010.
\newblock \href {https://doi.org/10.1016/j.wocn.2008.10.004} {Quantal theory,
  enhancement and overlap}.
\newblock \emph{Journal of Phonetics}, 38(1):10--19.

\bibitem[{Stolcke(2002)}]{stolcke2002srilm}
Andreas Stolcke. 2002.
\newblock \href
  {https://www.isca-speech.org/archive/archive_papers/icslp_2002/i02_0901.pdf}
  {{SRILM} - an extensible language modeling toolkit}.
\newblock In \emph{Seventh International Conference on Spoken Language
  Processing}, pages 901--904.

\bibitem[{Vaux and Samuels(2015)}]{VauxSamuels2015}
Bert Vaux and Bridget Samuels. 2015.
\newblock \href {https://doi.org/10.1515/tlr-2014-0028} {Explaining vowel
  systems: Dispersion theory vs natural selection}.
\newblock \emph{Linguistic Review}, 32(3):573--599.

\bibitem[{Watt(2000)}]{Watt2000}
Dominic J.~L. Watt. 2000.
\newblock \href {https://doi.org/10.1017/S0954394500121040} {Phonetic parallels
  between the close-mid vowels of {T}yneside {E}nglish: {A}re they internally
  or externally motivated?}
\newblock \emph{Language Variation and Change}, 12(1):69--101.

\bibitem[{Wells(1995/2000)}]{wells1995computer}
John~C. Wells. 1995/2000.
\newblock \href {https://www.phon.ucl.ac.uk/home/sampa/x-sampa.htm}
  {Computer-coding the {IPA}: A proposed extension of {SAMPA}}.

\bibitem[{Whalen and Levitt(1995)}]{WhalenLevitt1995}
D~.H. Whalen and Andrea~G. Levitt. 1995.
\newblock \href
  {https://www.sciencedirect.com/science/article/abs/pii/S0095447095801650}
  {The universality of intrinsic {F}0 of vowels}.
\newblock \emph{Journal of Phonetics}, 23:349--366.

\bibitem[{Wiesner et~al.(2019)Wiesner, Adams, Yarowsky, Trmal, and
  Khudanpur}]{wiesner2019asru}
Matthew Wiesner, Oliver Adams, David Yarowsky, Jan Trmal, and Sanjeev
  Khudanpur. 2019.
\newblock \href {https://ieeexplore.ieee.org/document/9004019} {Zero-shot
  pronunciation lexicons for cross-language acoustic model transfer}.
\newblock In \emph{Proceedings of IEEE Association for Automatic Speech
  Recognition and Understanding (ASRU)}.

\bibitem[{Zhang et~al.(2017)Zhang, Manohar, Povey, and
  Khudanpur}]{zhang2017acoustic}
Xiaohui Zhang, Vimal Manohar, Daniel Povey, and Sanjeev Khudanpur. 2017.
\newblock \href {https://arxiv.org/abs/1706.03747} {Acoustic data-driven
  lexicon learning based on a greedy pronunciation selection framework}.
\newblock \emph{arXiv preprint arXiv:1706.03747}.

\bibitem[{Zwicker and Terhardt(1980)}]{ZwickerTerhardt1980}
Eberhard Zwicker and Ernst Terhardt. 1980.
\newblock \href {https://doi.org/10.1121/1.385079} {Analytical expressions for
  critical-band rate and critical bandwidth as a function of frequency}.
\newblock \emph{The Journal of the Acoustical Society of America},
  68(5):1523--1525.

\end{thebibliography}
\bibliographystyle{acl_natbib}

\newpage
%-------------------------
\appendix
\label{sec:appendix}
%------------------
% \renewcommand{\thepage}{} %stops page counter for appendices which is nice for submission where they'll be 2 pdfs
%------------------

%-------------------------
\newpage\onecolumn
%-----------------------

%------------------------------------------------------------
\section{Pairwise Correlations between Vowel Formant Measures ~(\cref{sec:case-studies} Case Studies)}\label{app:cors}

\cref{tab:f1-cors} and \cref{tab:f2-cors} respectively show Pearson correlations of mean F1 and mean F2 in ERB between vowels that appear in at least 10 readings. As formalized in the present analysis, phonetic uniformity predicts strong correlations of mean F1 among vowels with a shared height specification, and strong correlations of mean F2 among vowels with a shared backness specification. The respective ``Height'' and ``Backness'' columns in \cref{tab:f1-cors} and \cref{tab:f2-cors} indicate whether the vowels in each pair match in their respective specifications. $p$-values are corrected for multiple comparisons using the Benjamini-Hochberg correction and a false discovery rate of 0.25 \cite{benjamini1995controlling}. Significance is assessed at $\alpha$ = 0.05 following the correction for multiple comparisons; rows that appear in gray have correlations that are not significant according to this threshold.

%-----------------
\begin{table}[h]
\centering
\footnotesize
\begin{minipage}{.45\linewidth}
\centering
\setlength\tabcolsep{5pt} 
\begin{tabular}{llcccc}
\midrule[\heavyrulewidth]
V1 & V2 & Height & \# Readings & $r$ & $p$ \\ 
\midrule[\heavyrulewidth]
\phon{i} & \phon{i:} & \checkmark & 12 & 0.81 & 0.006 \\
\phon{e:} & \phon{o:} & \checkmark & 10 & 0.81 & 0.015 \\
\phon{i} & \phon{u} & \checkmark & 40 & 0.79 & 0.000 \\
\rowcolor[HTML]{dcdcdc}
\phon{E} & \phon{O} & \checkmark & 11 & 0.68 & 0.053 \\
\phon{o} & \phon{a} &  & 37 & 0.66 & 0.000 \\
\rowcolor[HTML]{dcdcdc}
\phon{i:} & \phon{o:} &  & 11 & 0.65 & 0.070 \\
\rowcolor[HTML]{dcdcdc}
\phon{i:} & \phon{u:} & \checkmark & 12 & 0.64 & 0.061 \\
\phon{e} & \phon{o} & \checkmark & 35 & 0.62 & 0.001 \\
\phon{e} & \phon{u} &  & 36 & 0.59 & 0.001 \\
\phon{e} & \phon{a} &  & 34 & 0.58 & 0.002 \\
\rowcolor[HTML]{dcdcdc}
\phon{u} & \phon{@} &  & 12 & 0.58 & 0.105 \\
\rowcolor[HTML]{dcdcdc}
\phon{i:} & \phon{e:} &  & 11 & 0.58 & 0.118 \\
\phon{i} & \phon{e} &  & 38 & 0.54 & 0.002 \\
\rowcolor[HTML]{dcdcdc}
\phon{E} & \phon{a} &  & 12 & 0.54 & 0.127 \\
\phon{u} & \phon{o} &  & 38 & 0.49 & 0.007 \\
\rowcolor[HTML]{dcdcdc}
\phon{E} & \phon{u} &  & 14 & 0.49 & 0.135 \\
\phon{i} & \phon{o} &  & 39 & 0.46 & 0.011 \\
\rowcolor[HTML]{dcdcdc}
\phon{e} & \phon{E} & \checkmark & 12 & 0.46 & 0.204 \\
\phon{u} & \phon{a} &  & 37 & 0.42 & 0.027 \\
\rowcolor[HTML]{dcdcdc}
\phon{i:} & \phon{e} &  & 11 & 0.42 & 0.288 \\
\rowcolor[HTML]{dcdcdc}
\phon{u} & \phon{u:} & \checkmark & 10 & 0.41 & 0.334 \\
\rowcolor[HTML]{dcdcdc}
\phon{i:} & \phon{u} & \checkmark & 11 & 0.33 & 0.430 \\
\rowcolor[HTML]{dcdcdc}
\phon{i:} & \phon{a} &  & 11 & 0.28 & 0.496 \\
\rowcolor[HTML]{dcdcdc}
\phon{i} & \phon{a} &  & 39 & 0.27 & 0.173 \\
\rowcolor[HTML]{dcdcdc}
\phon{i} & \phon{E} &  & 14 & 0.24 & 0.496 \\
\rowcolor[HTML]{dcdcdc}
\phon{i:} & \phon{o} &  & 13 & 0.19 & 0.624 \\
\rowcolor[HTML]{dcdcdc}
\phon{i} & \phon{@} &  & 13 & 0.10 & 0.785 \\
\rowcolor[HTML]{dcdcdc}
\phon{u} & \phon{O} &  & 12 & 0.09 & 0.785 \\
\rowcolor[HTML]{dcdcdc}
\phon{E} & \phon{o} & \checkmark & 13 & -0.09 & 0.785 \\
\rowcolor[HTML]{dcdcdc}
\phon{e} & \phon{O} & \checkmark & 10 & -0.12 & 0.785 \\
\rowcolor[HTML]{dcdcdc}
\phon{u:} & \phon{o} &  & 10 & -0.12 & 0.785 \\
\rowcolor[HTML]{dcdcdc}
\phon{i} & \phon{O} &  & 11 & -0.42 & 0.288 \\
\rowcolor[HTML]{dcdcdc}
\phon{o} & \phon{@} & \checkmark & 11 & -0.51 & 0.173 \\
\phon{@} & \phon{a} &  & 11 & -0.90 & 0.001 \\
\bottomrule
\end{tabular}
\caption{Pearson correlations ($r$) of \textbf{mean F1} in ERB between vowel categories.}
\label{tab:f1-cors}
\end{minipage}
\hfill
\begin{minipage}{.45\linewidth}
\centering
\footnotesize
\setlength\tabcolsep{5pt} 
\begin{tabular}{llcccc}
\midrule[\heavyrulewidth]
 V1 & V2 & Backness & \# Readings & $r$ & $p$ \\ 
\midrule[\heavyrulewidth]
\phon{e} & \phon{E} & \checkmark & 12 & 0.77 & 0.019 \\
\phon{u} & \phon{u:} & \checkmark & 10 & 0.77 & 0.037 \\
\phon{i} & \phon{i:} & \checkmark & 12 & 0.70 & 0.038 \\
\phon{u} & \phon{o} & \checkmark & 38 & 0.69 & 0.000 \\
\phon{i} & \phon{E} & \checkmark & 14 & 0.69 & 0.031 \\
\rowcolor[HTML]{dcdcdc}
\phon{u:} & \phon{o} & \checkmark & 10 & 0.62 & 0.130 \\
\rowcolor[HTML]{dcdcdc}
\phon{u} & \phon{@} &  & 12 & 0.60 & 0.107 \\
\rowcolor[HTML]{dcdcdc}
\phon{u} & \phon{O} & \checkmark & 12 & 0.52 & 0.168 \\
\phon{i} & \phon{e} & \checkmark & 38 & 0.41 & 0.038 \\
\rowcolor[HTML]{dcdcdc}
\phon{E} & \phon{a} &  & 12 & 0.32 & 0.519 \\
\rowcolor[HTML]{dcdcdc}
\phon{o} & \phon{a} &  & 37 & 0.30 & 0.159 \\
\rowcolor[HTML]{dcdcdc}
\phon{e:} & \phon{o:} &  & 10 & 0.27 & 0.666 \\
\rowcolor[HTML]{dcdcdc}
\phon{e} & \phon{a} &  & 34 & 0.24 & 0.339 \\
\rowcolor[HTML]{dcdcdc}
\phon{o} & \phon{@} &  & 11 & 0.21 & 0.724 \\
\rowcolor[HTML]{dcdcdc}
\phon{@} & \phon{a} & \checkmark & 11 & 0.16 & 0.830 \\
\rowcolor[HTML]{dcdcdc}
\phon{i:} & \phon{e} & \checkmark & 11 & 0.11 & 0.911 \\
\rowcolor[HTML]{dcdcdc}
\phon{i} & \phon{a} &  & 39 & 0.06 & 0.911 \\
\rowcolor[HTML]{dcdcdc}
\phon{i:} & \phon{e:} & \checkmark & 11 & 0.06 & 0.965 \\
\rowcolor[HTML]{dcdcdc}
\phon{e} & \phon{o} &  & 35 & 0.01 & 0.965 \\
\rowcolor[HTML]{dcdcdc}
\phon{u} & \phon{a} &  & 37 & 0.00 & 0.985 \\
\rowcolor[HTML]{dcdcdc}
\phon{E} & \phon{O} &  & 11 & -0.03 & 0.965 \\
\rowcolor[HTML]{dcdcdc}
\phon{i:} & \phon{a} &  & 11 & -0.04 & 0.965 \\
\rowcolor[HTML]{dcdcdc}
\phon{E} & \phon{o} &  & 13 & -0.04 & 0.965 \\
\rowcolor[HTML]{dcdcdc}
\phon{e} & \phon{u} &  & 36 & -0.12 & 0.666 \\
\rowcolor[HTML]{dcdcdc}
\phon{E} & \phon{u} &  & 14 & -0.22 & 0.666 \\
\rowcolor[HTML]{dcdcdc}
\phon{i} & \phon{@} &  & 13 & -0.23 & 0.666 \\
\rowcolor[HTML]{dcdcdc}
\phon{i:} & \phon{o:} &  & 11 & -0.42 & 0.345 \\
\phon{i} & \phon{o} &  & 39 & -0.48 & 0.017 \\
\rowcolor[HTML]{dcdcdc}
\phon{i:} & \phon{o} &  & 13 & -0.52 & 0.149 \\
\phon{i} & \phon{u} &  & 40 & -0.55 & 0.003 \\
\rowcolor[HTML]{dcdcdc}
\phon{i} & \phon{O} &  & 11 & -0.63 & 0.107 \\
\rowcolor[HTML]{dcdcdc}
\phon{e} & \phon{O} &  & 10 & -0.65 & 0.107 \\
\phon{i:} & \phon{u} &  & 11 & -0.80 & 0.019 \\
\phon{i:} & \phon{u:} &  & 12 & -0.83 & 0.009 \\
\bottomrule
\end{tabular}
\caption{Pearson correlations ($r$) of \textbf{mean F2} in ERB between vowel categories.}
\label{tab:f2-cors}
\end{minipage}
\end{table}

%-------------------------------------------------------------
\vspace{2em}
\section{Distributions of Unitran Segment Accuracy ~(\cref{sec:qualitymeasures} Quality Measures)}
\label{app:ali_hists}

Here we evaluate the quality of the Unitran dataset in more detail.  The goal is to explore the variation in the quality of the labeled Unitran segments across different languages and phoneme labels.  This evaluation includes only readings in high-resource languages, where we have not only the aligned Unitran pronunciations but also aligned high-resource pronunciations (Epitran or WikiPron) against which to evaluate them.  
The per-token statistics used to calculate these plots are included in the corpus release to enable closer investigation of individual phonemes than is possible here.

\subsection{Unitran Pronunciation Accuracy}

\begin{figure}[ht]
 \centering
 \includegraphics[width=0.9\textwidth]{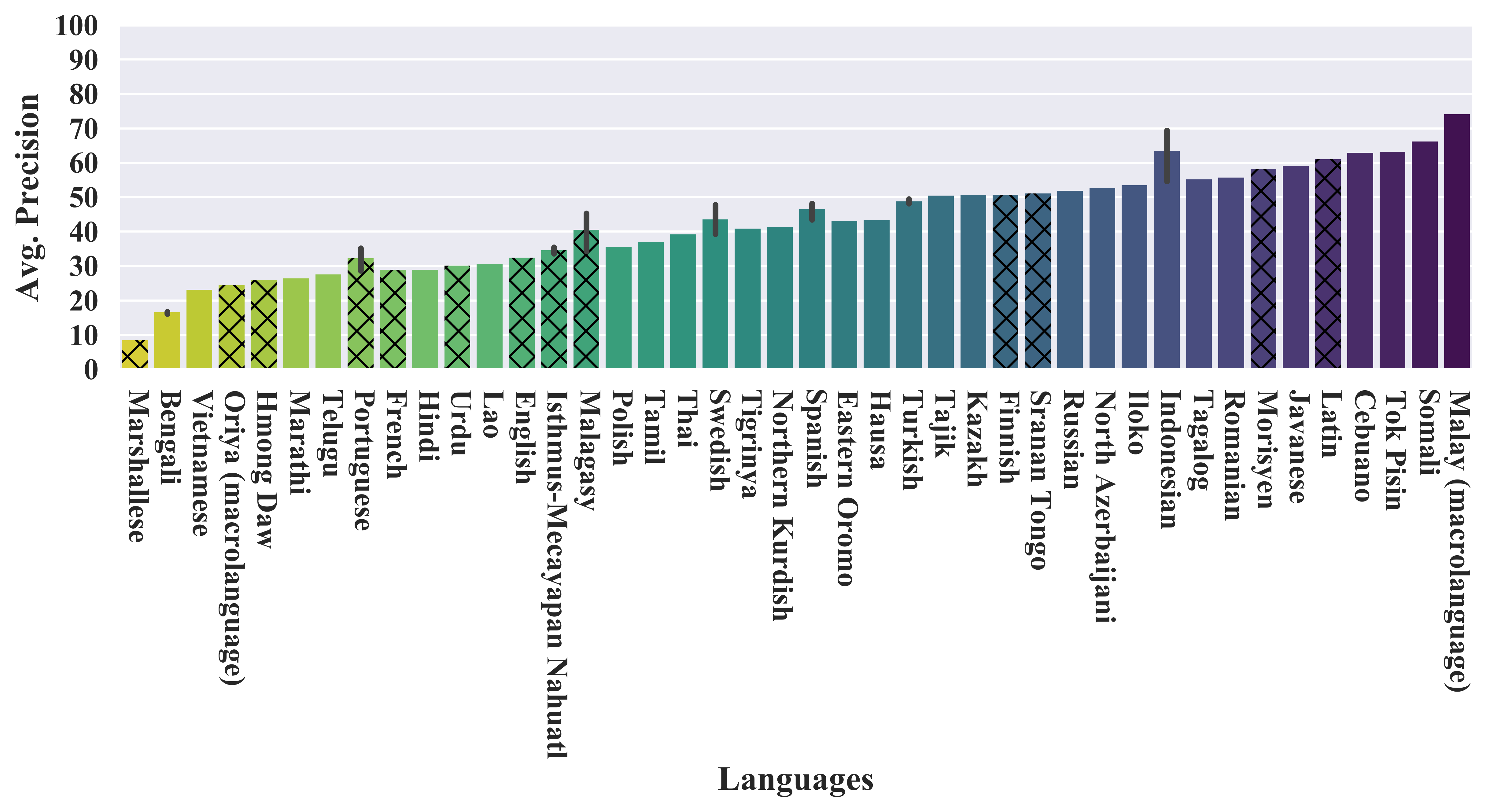}
 \caption{\textbf{Unitran pronunciation accuracy per language}, evaluated by Levenshtein alignment to WikiPron pronunciations (hatched bars) or Epitran pronunciations (plain bars).  
 Where a language has multiple readings, error bars show the min and max across those readings.}
 \label{fig:lngavg-acc-hist}
\end{figure}

\begin{figure}[ht]
 \centering
 \includegraphics[width=0.9\textwidth]{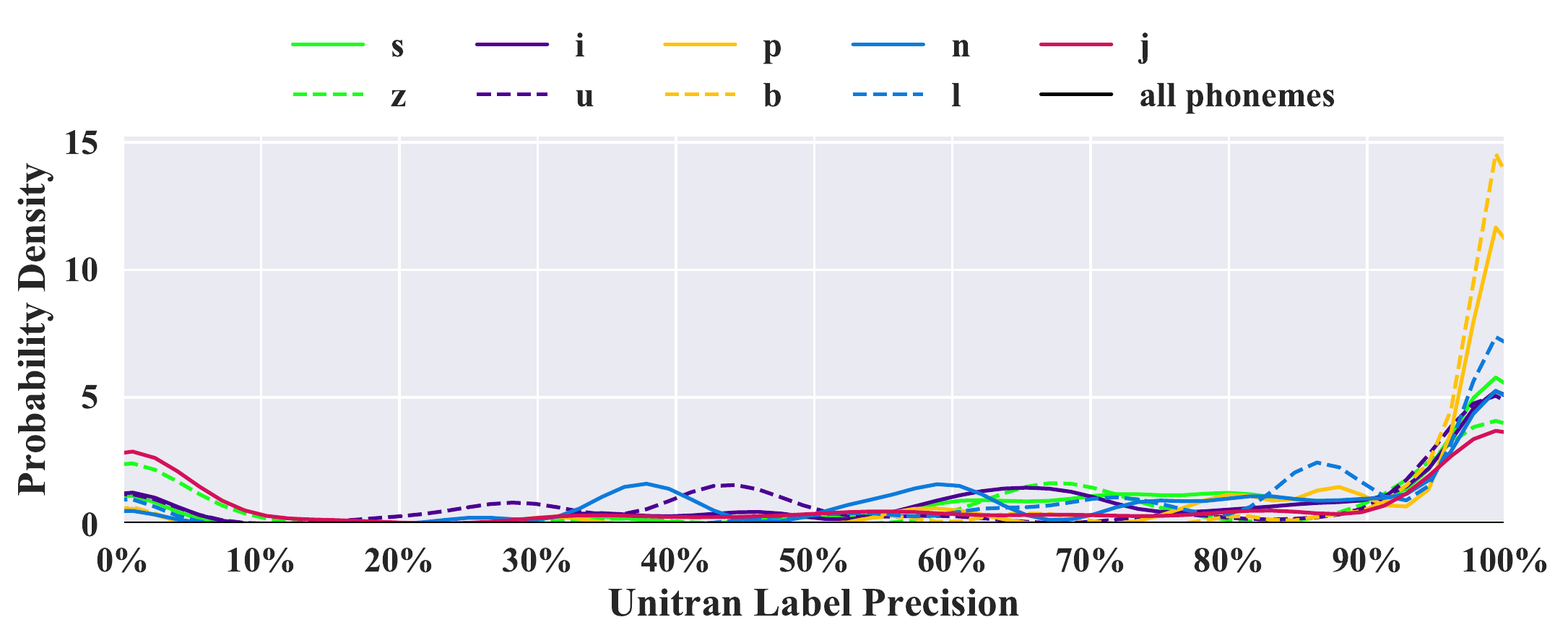}
 \caption{\textbf{Unitran pronunciation accuracy per language, for selected phonemes.} Accuracy is evaluated by Levenshtein alignment as in \cref{fig:lngavg-acc-hist}.  Each curve is a kernel density plot with integral 1.  For the \phon{z} curve, the integral between 80\% and 100\% (for example) is the estimated probability that in a high-resource language drawn uniformly at random, the fraction of Unitran \phon{z} segments that align to high-resource \phon{z} segments falls in that range.  The `all' curve is the same, but now the uniform draw is from all pairs of (high-resource language, Unitran phoneme used in that language).}
 \label{fig:lngphn-acc-hist}
\end{figure}

First, in \cref{fig:lngavg-acc-hist,fig:lngphn-acc-hist}, we consider whether Unitran's utterance pronunciations are accurate without looking at the audio.  For each utterance, we compute the unweighted Levenshtein alignment between the Unitran pronunciation of the utterance and the high-resource pronunciation.  For each reading, we then score the percentage of Unitran `phoneme' tokens that were aligned to high-resource `phoneme' tokens with exactly the same label.\footnote{By contrast, PER in \cref{tab:PER} aligns at the word level rather than the utterance level, uses the number of symmetric alignment errors (insertions + deletions + substitutions) rather than the number of correct Unitran phonemes, and normalizes by the length of the high-resource `reference' pronunciation rather than by the length of the Unitran pronunciation.}
We can see in \cref{fig:lngphn-acc-hist} that many labels are highly accurate in many readings while being highly inaccurate in many others.  Some labels are noisy in some readings.\footnote{Note that as \cref{sec:qualitymeasures} points out, it may be unfair to require exact match of labels, since annotation schemes vary.)}

\newpage
\subsection{Unitran Segment Label Accuracy}

\begin{figure}[ht]
 \centering
  \includegraphics[width=0.92\textwidth]{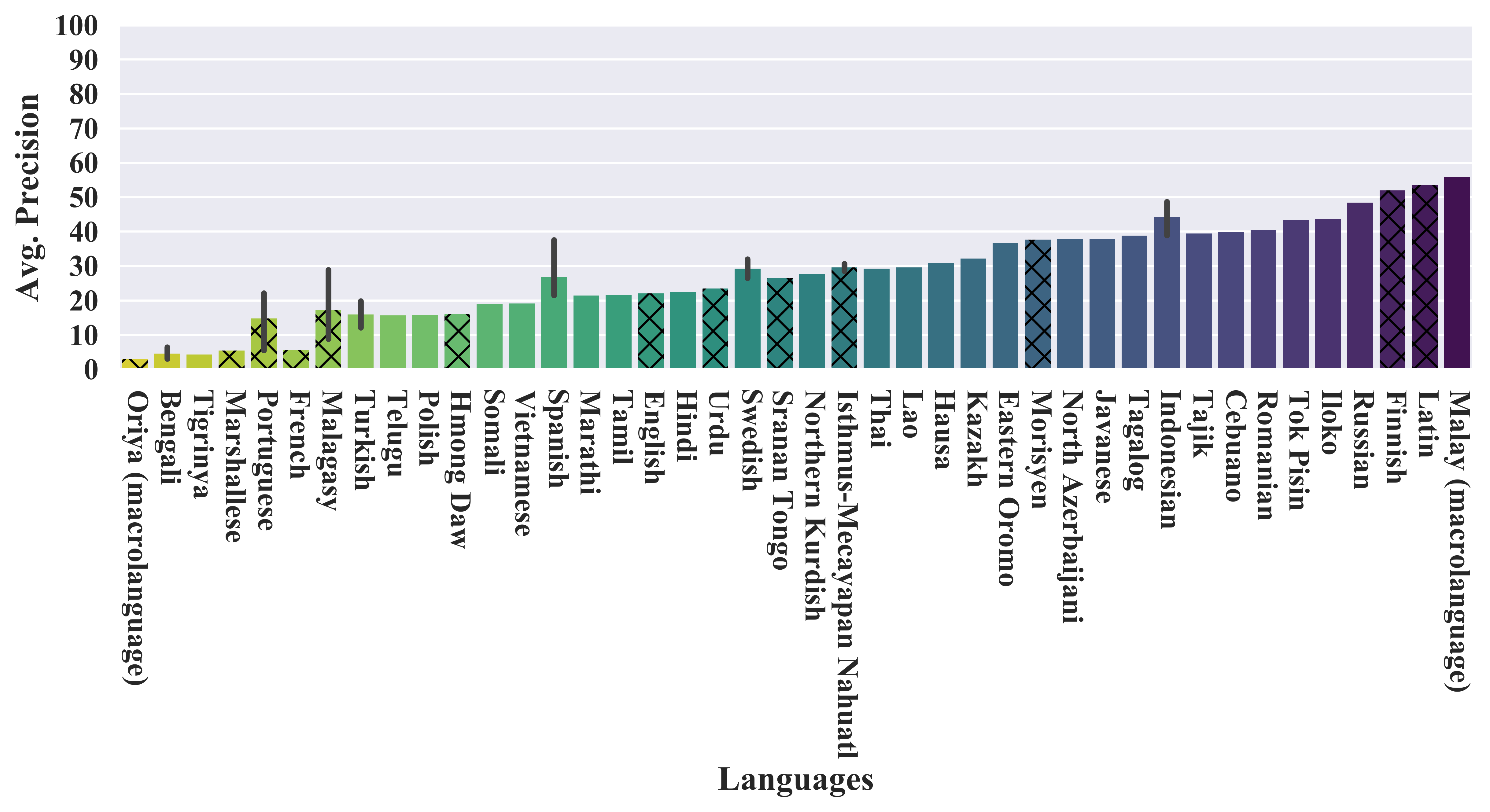}
 \caption{\textbf{Unitran pronunciation accuracy per language}, as in \cref{fig:lngavg-acc-hist} but with audio midpoint alignment in place of Levenshtein alignment.}
  \label{fig:lngavg-mid-hist}
\end{figure}

\begin{figure}[ht]
 \centering
  \includegraphics[width=0.92\textwidth]{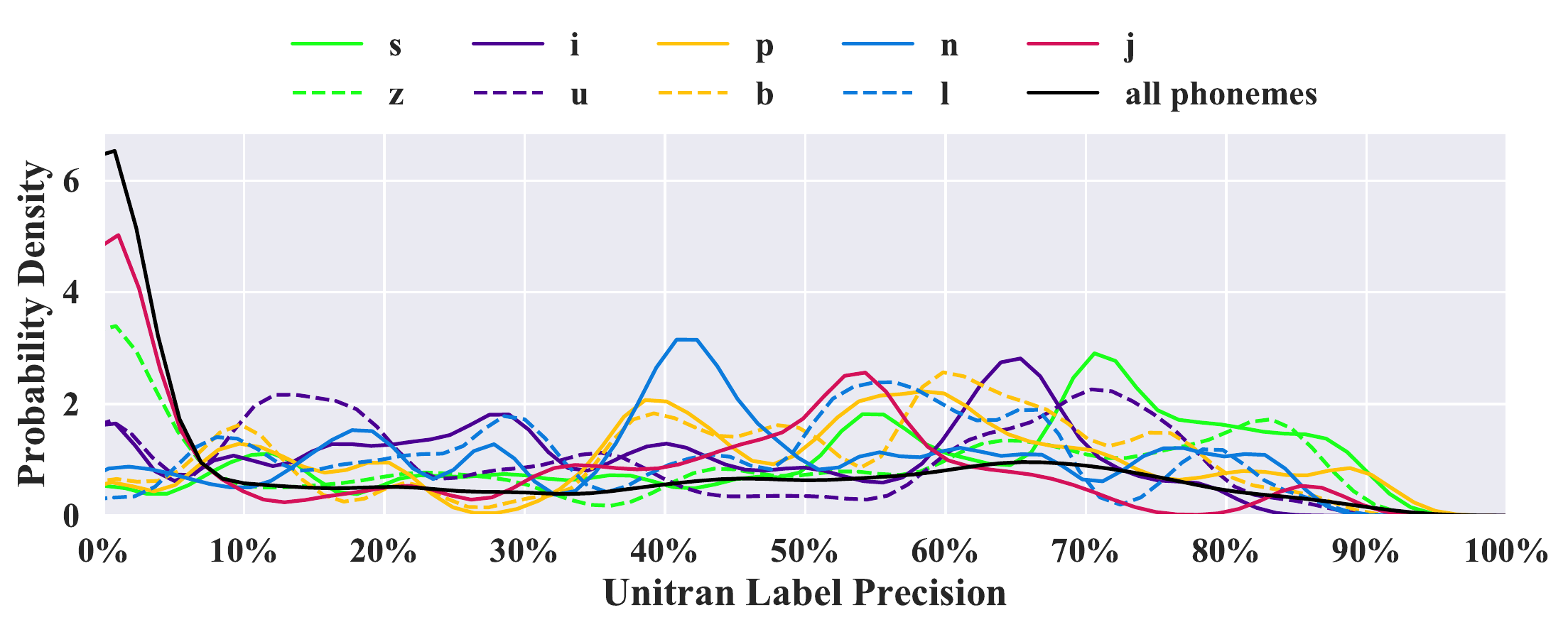}
 \caption{\textbf{Unitran pronunciation accuracy per language, for selected phonemes,} as in \cref{fig:lngphn-acc-hist} but with audio midpoint alignment in place of Levenshtein alignment.}
  \label{fig:lngphn-mid-hist}
\end{figure}

In \cref{fig:lngavg-mid-hist,fig:lngphn-mid-hist}, we ask the same question again, but making use of the audio data.  The match for each Unitran segment is now found not by Levenshtein alignment, but more usefully by choosing the high-resource segment with the closest midpoint.  For each reading, we again score the percentage of Unitran `phoneme' tokens whose aligned high-resource `phoneme' tokens have exactly the same label.  Notice that phonemes that typically had high accuracy in \cref{fig:lngphn-acc-hist}, such as \phon{p} and \phon{b}, now have far more variable accuracy in \cref{fig:lngphn-mid-hist}, suggesting difficulty in aligning the Unitran pronunciations to the correct parts of the audio.

\newpage
\subsection{Unitran Segment Boundary Accuracy}

\begin{figure}[ht]
 \centering
 \includegraphics[width=0.88\textwidth]{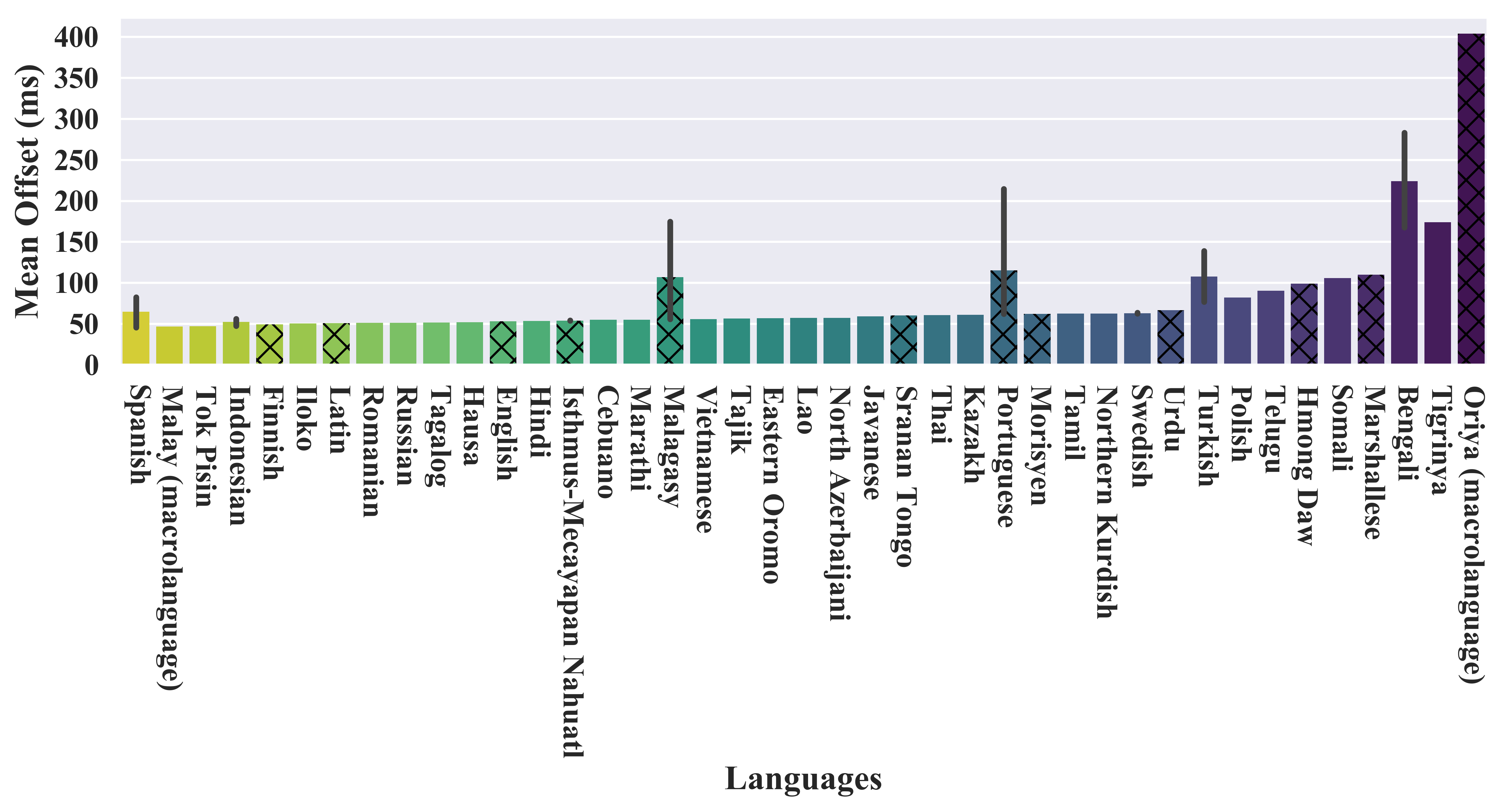}
 \caption{\textbf{Mean error per language in the temporal boundaries of Unitran segments.}. Each Unitran segment is evaluated against the WikiPron segment (hatched bars) or Epitran segment (plain bars) with the closest midpoint, as if the latter were truth.  The error of a segment is the absolute offset of the left boundary plus the absolute offset of the right boundary.  Only segments where the Unitran label matches the Epitran/WikiPron label are included in the average.  Where a language has multiple readings, error bars show the min and max across those readings.}
 \label{fig:lngavg-bound-acc-hist}
\end{figure}

\begin{figure}[h]
 \centering
 \includegraphics[width=0.88\textwidth]{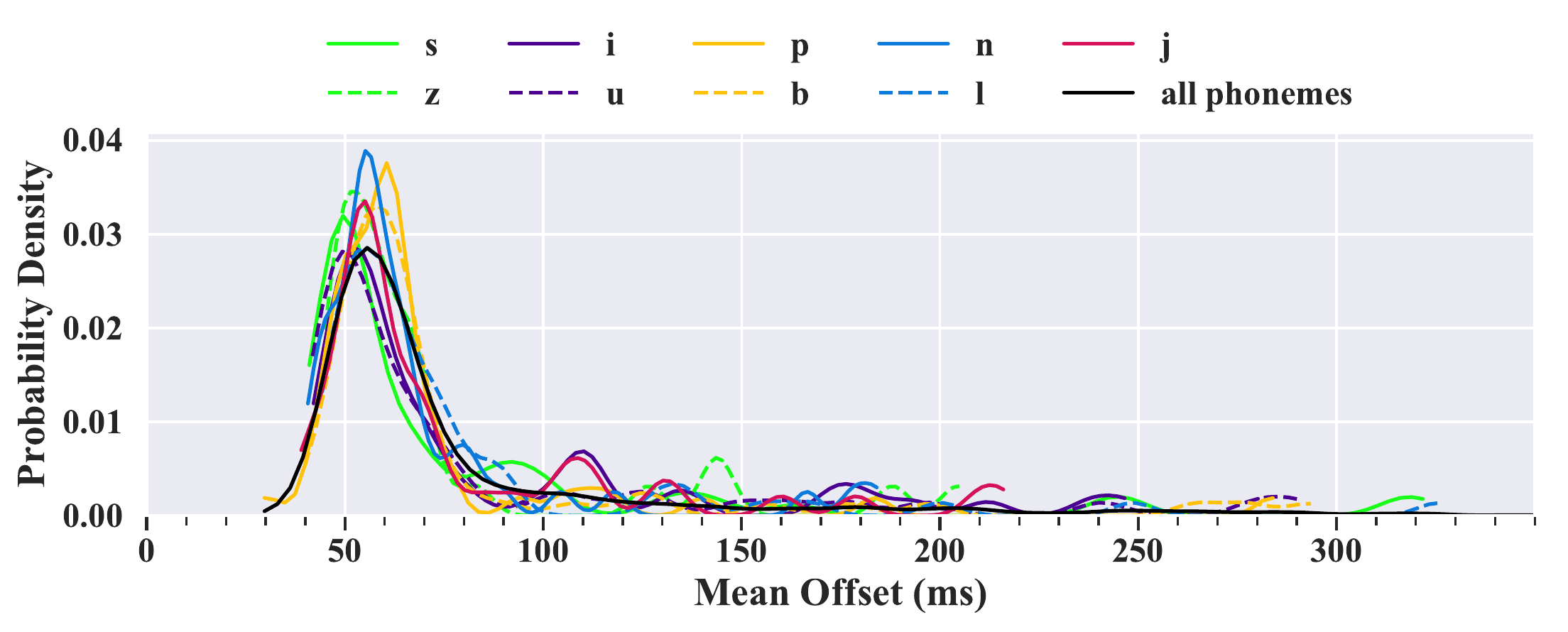}
 \caption{\textbf{Mean error per language in the temporal boundaries of Unitran segments, for selected phonemes.}  Each curve is a kernel density plot with integral 1.  For the \phon{z} curve, the integral between 50ms and 100ms (for example) is the estimated probability that in a high-resource language drawn uniformly at random, the Unitran \phon{z} segments whose corresponding Epitran or WikiPron segments are also labeled with \phon{z} have mean boundary error in that range.  Small bumps toward the right correspond to individual languages where the mean error of \phon{z} is unusually high.
 The `all' curve is the same, but now the uniform draw is from all pairs of (high-resource language, Unitran phoneme used in that language).  The boundary error of a segment is evaluated as in  \cref{fig:lngavg-bound-acc-hist}.}
 \label{fig:lngphn-bound-acc-hist}
\end{figure}

Finally, in \cref{fig:lngavg-bound-acc-hist,fig:lngphn-bound-acc-hist}, we measure whether Unitran segments with the ``correct'' label also have the ``correct'' time boundaries, where ``correctness'' is evaluated against the corresponding segments obtained using Epitran or WikiPron+G2P.  

%------------------------------------------------------------
\newpage\clearpage
\section{WikiPron Grapheme-to-Phoneme (G2P) Accuracy ~(\cref{sec:qualitymeasures} Quality Measures)}
\label{app:wikipron_acc}

For each language where we used WikiPron, 
\cref{tab:wiki-g2p} shows the phoneme error rate (PER) of Phonetisaurus G2P models trained on WikiPron entries, as evaluated on held-out WikiPron entries.  This is an estimate of how accurate our G2P-predicted pronunciations are on out-of-vocabulary words, insofar as those are distributed similarly to the in-vocabulary words.  (It is possible, however, that out-of-vocabulary words such as Biblical names are systematically easier or harder for the G2P system to pronounce, depending on how they were transliterated.)

The same G2P configuration was used for all languages, with the hyperparameter settings shown in \cref{tab:g2p_hyperparams}.  (\texttt{seq1\_max} and \texttt{seq2\_max} describe how many tokens in the grapheme and phoneme sequences can align to each other.). These settings were tuned on SIGMORPHON 2020 Task 1 French, Hungarian, and Korean data \citep{gorman2020sigmorphon}, using 20 random 80/20 splits.

\vspace{1em}
\begin{table}[h]
\begin{adjustbox}{width=\linewidth}
\begin{tabular}{lrrrrrrrrrrrrrrr}
\toprule
\textbf{ISO 639-3} & \textbf{fin} & \textbf{lat} & \textbf{nhx} & \textbf{srn} & \textbf{mah} & \textbf{por-po} & \textbf{mfe} & \textbf{mww} & \textbf{por-bz} & \textbf{eng} & \textbf{khm} & \textbf{mlg} & \textbf{ori} & \textbf{ban} & \textbf{urd} \\
% Test size          & 8348         & 10917        & 6836         & 25           & 31           & 162          & 1926            & 40           & 45           & 2015            & 10860        & 603          & 22           & 42           & 34           & 140          \\
Train size         & 41741         & 34181        & 126          & 157          & 813          & 9633            & 203          & 227          & 10077           & 54300           & 3016         & 114          & 211          & 172          & 704          \\
\cmidrule(lr){2-2} \cmidrule(lr){3-3} \cmidrule(lr){4-4} \cmidrule(lr){5-5} \cmidrule(lr){6-6} \cmidrule(lr){7-7} \cmidrule(lr){8-8} \cmidrule(lr){9-9} \cmidrule(lr){10-10} \cmidrule(lr){11-11} \cmidrule(lr){12-12} \cmidrule(lr){13-13} \cmidrule(lr){14-14} \cmidrule(lr){15-15} \cmidrule(lr){16-16} 
%-----------
\textbf{PER}       & 0.8       & 2.4          & 4.1          & 4.6          & 9.6          & 10.1            & 10.7         & 10.8         & 11.4            & 14.5         & 15.5         & 15.8         & 16.1         & 19.5         & 26.7         \\
%-----------
        & $\pm$0.02     & $\pm$0.04     & $\pm$1.02     & $\pm$0.76     & $\pm$0.41    & $\pm$0.11        & $\pm$1.2     & $\pm$1.29     & $\pm$0.16        & $\pm$0.06    & $\pm$0.38     & $\pm$1.44     & $\pm$1.13     & $\pm$1.35     & $\pm$0.60    \\
\bottomrule
\end{tabular}
\end{adjustbox}
\caption{WikiPron G2P Phone Error Rate (PER) calculated treating WikiPron annotations as ground-truth. We perform 20 trials with random 80/20 splits per language, and report PER averaged across trials with 95\% confidence intervals for each language.} 
\label{tab:wiki-g2p}
\end{table}

\vspace{1em}
\begin{table}[h]
\centering
\begin{adjustbox}{width=\linewidth}
\begin{tabular}{c|ccccccc}
\toprule
\textbf{Phonetisaurus Alignment}& \texttt{seq1\_max} & \texttt{seq2\_max} & \texttt{seq1\_del}  & \texttt{seq2\_del} & \texttt{grow} & \texttt{max EM iterations} \\
\textbf{Hyperparameters} & 1 & 3 & True & True & True & 11 \\
\midrule
\textbf{Graphone Language Model} & \texttt{n-gram order} & \texttt{LM type} & \texttt{discounting} & \texttt{gt2min} & \texttt{gt3min} & \texttt{gt4min} & \texttt{gt5min} \\
\textbf{Hyperparameters} & 5 & max-ent & Kneser-Ney & 2 & 2 & 3 & 4 \\
 \bottomrule
\end{tabular}
\end{adjustbox}
\caption{Table of final G2P hyperparameter settings. Alignment parameters not listed here for \texttt{phonetisaurus-align} use the default values. The language model was trained using SRILM \cite{stolcke2002srilm} \texttt{ngram-count} using default values except for those listed above.}
\label{tab:g2p_hyperparams}
\end{table}

%-------------------------------------------------------------
\newpage\clearpage
\section{Retention Statistics ~(\cref{sec:data_filtering} Data Filtering) }
\label{app:retention}

\cref{tab:epi-uni-retention} shows what percentage of tokens would be retained after various methods are applied to filter out questionable tokens from the readings used in \cref{sec:data_filtering}.  In particular, the rightmost column shows the filtering that was actually used in \cref{sec:data_filtering}.  We compute statistics for each reading separately; in each column we report the minimum, median, mean, and maximum statistics over the readings. The top half of the table considers vowel tokens (for the vowels in \cref{app:cors}); the bottom half considers sibilant tokens (\phon{s} and \phon{z}). 

On the left side of the table, we consider three filtering techniques for Unitran alignments.  \textbf{Midpoint} retains only the segments whose labels are ``correct'' according to the midpoint-matching methods of \cref{app:ali_hists}.  \textbf{MCD} retains only those utterances with MCD $<$ 6.  \textbf{Outlier} removes tokens that are outliers according to the criteria described in \cref{sec:data_filtering}.  Finally, \textbf{AGG.} is the aggregate retention rate  retention rate after all three methods are applied in order. 

On the right side of the table, we consider the same filtering techniques for the high-resource alignments that we actually use, with the exception of \textbf{Midpoint}, as here we have no higher-quality annotation to match against.

\vspace{1em}
\begin{table}[h]
\centering
\footnotesize
\begin{tabular}{llrrrrr|rrrr}
 \toprule
 & & \multicolumn{4}{c}{\textbf{Unitran Alignments}} & & \multicolumn{4}{c}{\textbf{High-Resource Alignments}}\\
 \midrule
 & & \bf \# Tokens &  \bf Midpoint & \bf MCD & \bf Outlier & \bf AGG.        & \bf \# Tokens & \bf MCD & \bf Outlier & \bf AGG.\\ 
 \cmidrule(lr){3-3} \cmidrule(lr){4-4} \cmidrule(lr){5-5} \cmidrule(lr){6-6} \cmidrule(lr){7-7} \cmidrule(lr){8-8} \cmidrule(lr){9-9} \cmidrule(lr){10-10} \cmidrule(lr){11-11} 
\parbox[t]{2mm}{\multirow{5}{*}{\rotatebox[origin=c]{90}{\textit{Vowels}}}}
& Min               & 50,132            & 2\%      & 42\%  & 83\%    & 1\%    & 61,727     & 42\%  & 84\%    & 37\% \\
& Median            & 21,5162           & 23\%     & 88\%  & 90\%    & 16\%   & 232,059    & 88\%  & 90\%    & 79\%  \\
& Mean              & 23,9563           & 25\%     & 81\%  & 89\%    & 20\%   & 223,815    & 81\%  & 90\%    & 73\%  \\
& Max               & 662,813           & 65\%     & 100\% & 93\%    & 60\%   & 468,864    & 100\% & 93\%    & 93\%  \\
\cmidrule(lr){2-11}
& \# Readings       & 49               & 46       & 48  & 49      & 45     & 49        & 48    & 49      & 48   \\
\midrule
\parbox[t]{2mm}{\multirow{5}{*}{\rotatebox[origin=c]{90}{\textit{Sibilants}}}} 
& Min               & 7,198             & 10\%     & 42\%  & 89\%    & 13\%   & 7184      & 44\%  & 91\%    & 43\%  \\
& Median            & 28,690            & 70\%     & 87\%  & 97\%    & 59\%   & 27569     & 87\%  & 97\%    & 85\%  \\
& Mean              & 30,025            & 63\%     & 80\%  & 95\%    & 56\%   & 27083     & 81\%  & 96\%    & 79\%  \\
& Max               & 63,573            & 89\%     & 100\% & 98\%    & 79\%   & 45,290     & 100\% & 99\%    & 96\%  \\
\cmidrule(lr){2-11}
& \# Readings     & 36               & 26       & 35    & 36      & 19       & 25        & 22    & 25      & 22   \\
 \bottomrule
\end{tabular}
 \caption{Summary of quality measure retention statistics for \textbf{vowels} and \textbf{sibilants} over unique readings with reading-level MCD $<$ 8 for Unitran and high-resource alignments.}
\label{tab:epi-uni-retention}
\end{table}

%------------------------------------------------------------
\newpage\clearpage
\section{All \corpusname Languages} 
\label{app:lang_list}

All \numisos languages from \numrecs readings are presented here with their language family, ISO 639-3 code, and mean utterance alignment quality in Mel Cepstral Distortion (MCD) from \citet{black2019cmu}. 
Languages for which we release Epitran and/or WikiPron alignments in addition to Unitran alignments are marked with $^e$ and $^w$ respectively. 
MCD ranges from purple (\textit{low}), blue--green (\textit{mid}), to yellow (\textit{high}). 
Lower MCD typically corresponds to better audio-text utterance alignments and higher quality speech synthesis, but judgments regarding distinctions between languages may be subjective. 
ISO 639-3 is not intended to provide identifiers for \textit{dialects} or other \textit{sub-language} variations, 
which may be present here where there are multiple readings for one ISO 639-3 code. 
We report the most up-to-date language names from the ISO 639-3 schema \citep{ethnologue2020}. Language names and codes in many schema could be pejorative and outdated, but where language codes cannot be easily updated, language names can and often are.\looseness=-1

\begin{figure}[h]
 \centering
 \includegraphics[width=0.9\textwidth]{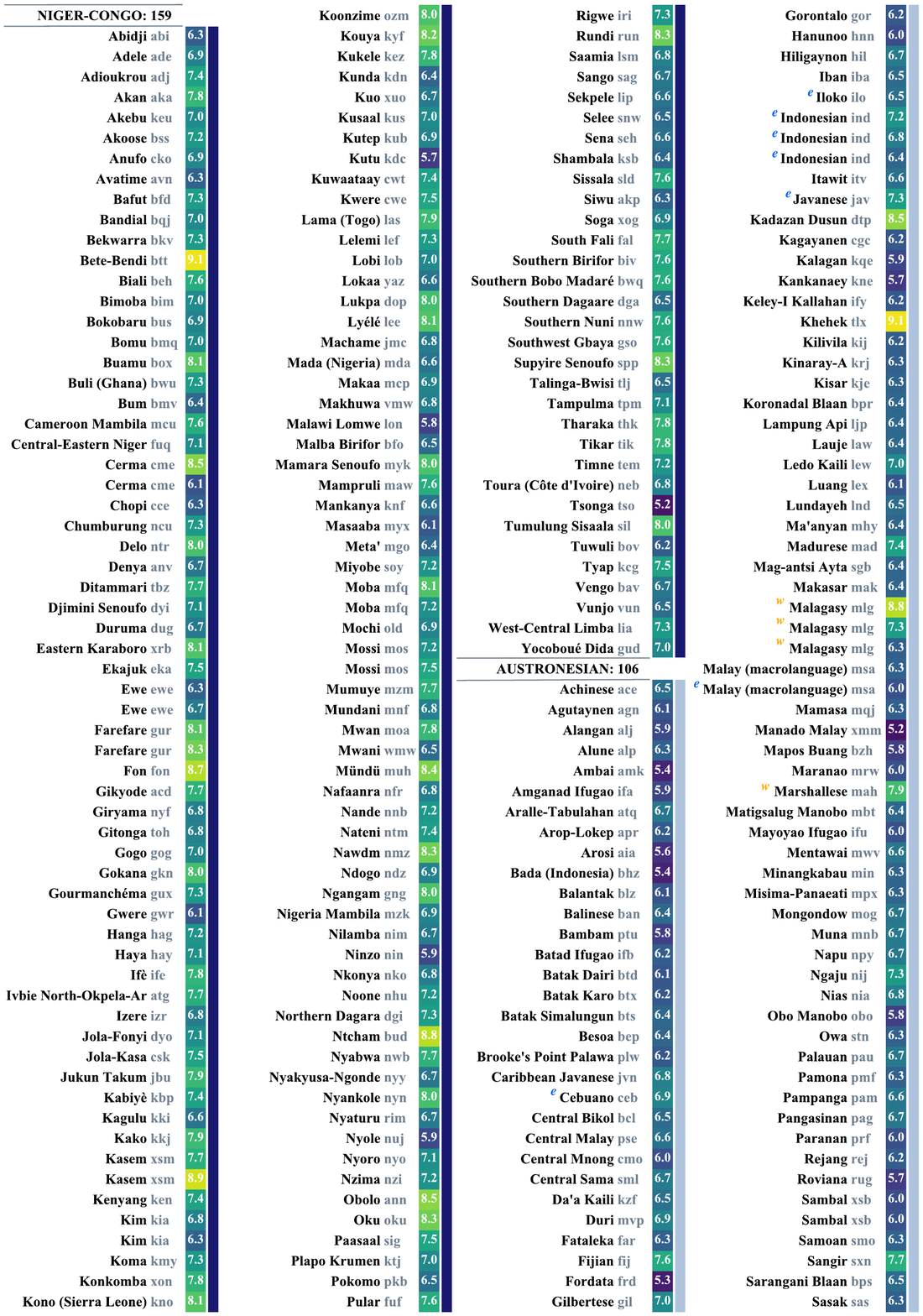}
  \label{fig:alllangs-part1}
\end{figure}

\begin{figure}[!h]
 \centering
 \includegraphics[width=\textwidth]{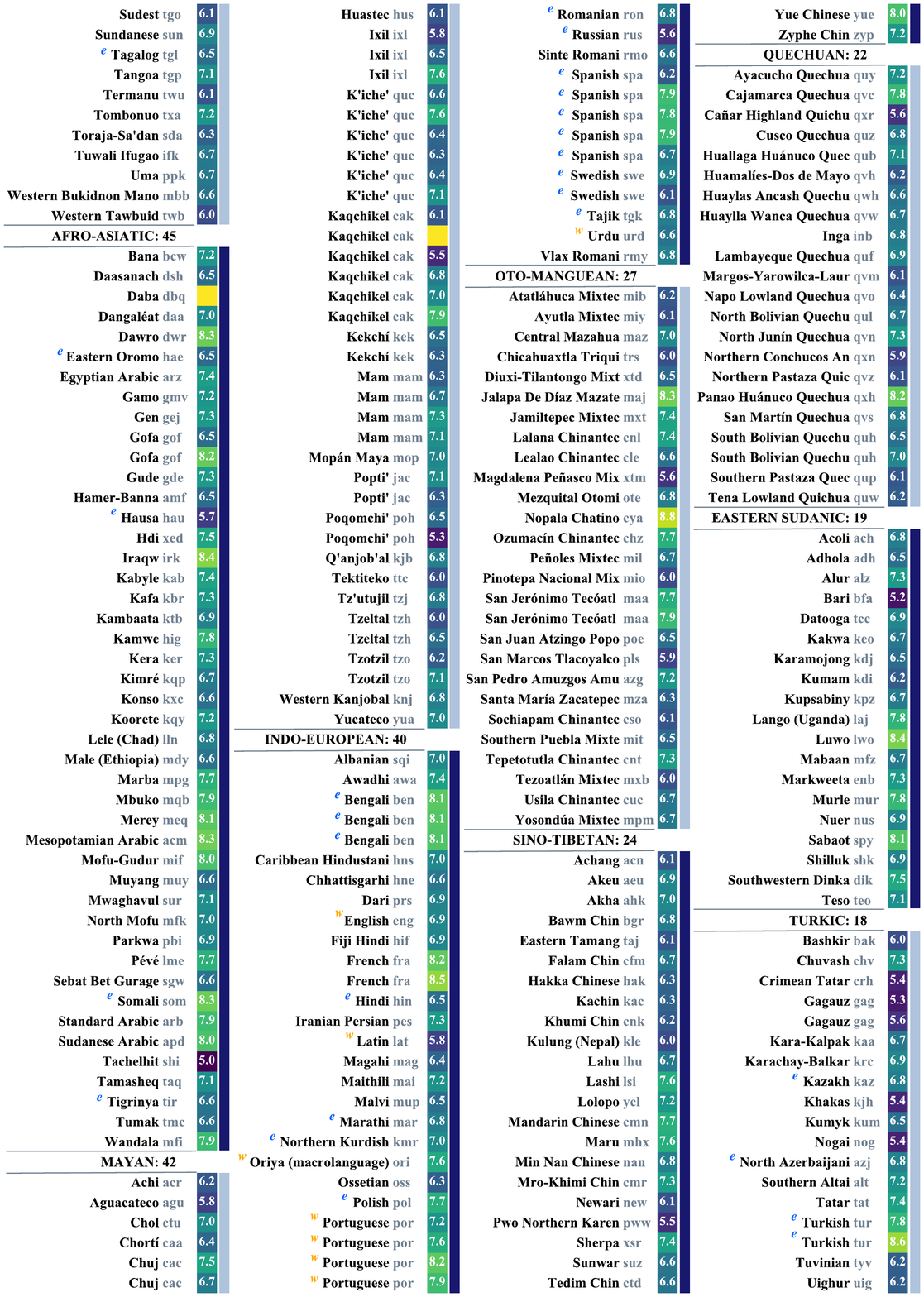}
  \label{fig:alllangs-part2}
\end{figure}

\begin{figure}[!h]
 \centering
 \includegraphics[width=\textwidth]{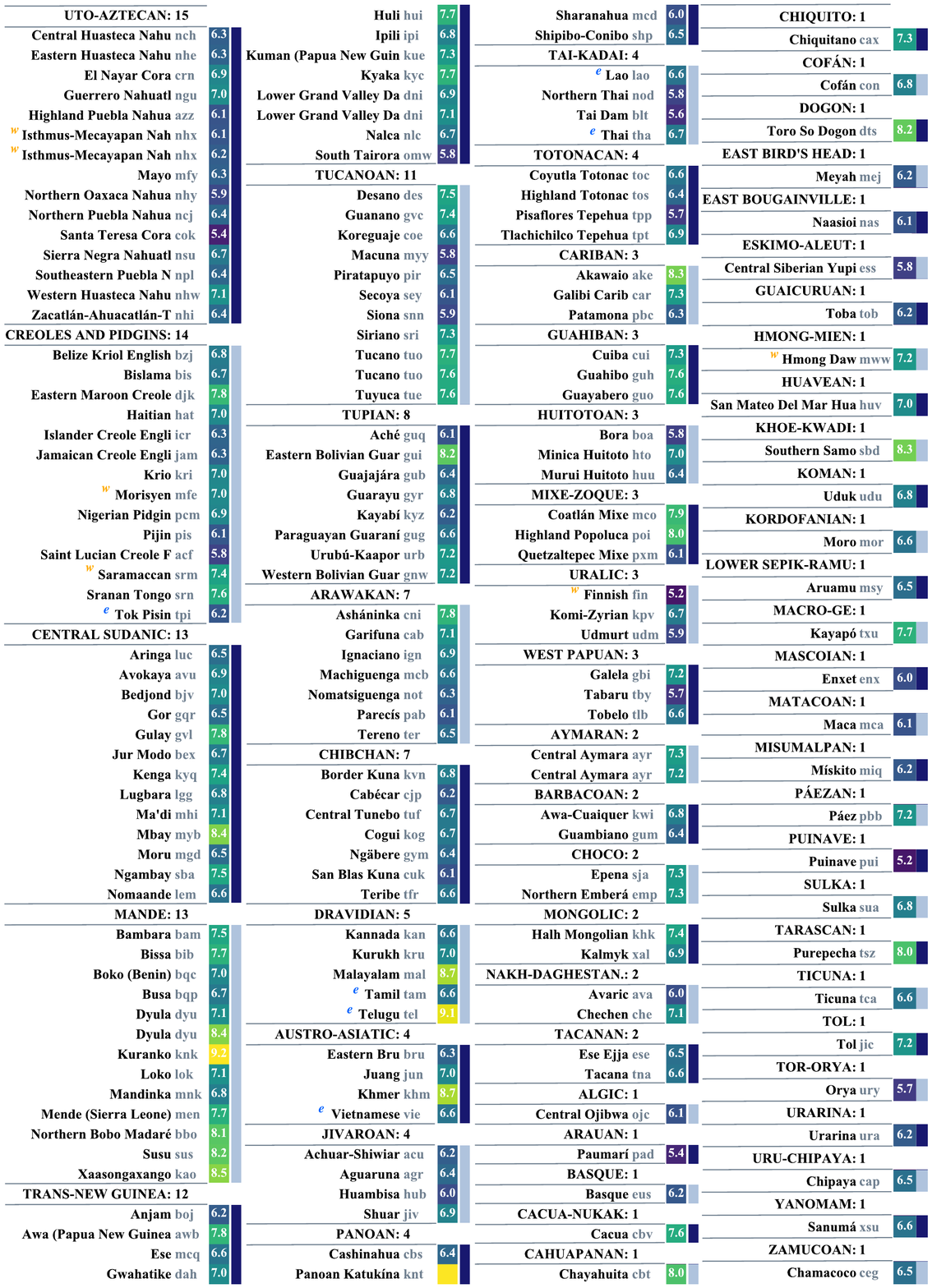}
  \label{fig:alllangs-part3}
\end{figure}

%------------------------------------------------------------
% ta ta for now!
%------------------------------------------------------------
\end{document}